\pdfoutput=1

\documentclass[11pt]{article}

\usepackage[final]{acl}

\usepackage{times}
\usepackage{latexsym}

\usepackage[T1]{fontenc}

\usepackage[utf8]{inputenc}

\usepackage{microtype}

\usepackage{inconsolata}

\usepackage{graphicx}

\usepackage{booktabs}
\usepackage{multirow}
\usepackage{float}
\usepackage{floatflt}
\usepackage{makecell}
\usepackage{array}
\usepackage{amsmath}
\usepackage{hyperref}
\usepackage{tabularx}
\usepackage{enumitem}
\usepackage{algorithmicx}
\usepackage{textcomp}
\usepackage{textcase} 
\usepackage{xcolor}
\usepackage{algorithm}
\usepackage{algpseudocode}
\usepackage{amssymb}
\usepackage{subcaption}
\usepackage{caption}
\usepackage{pifont}
\usepackage{array}
\usepackage{colortbl}
\usepackage{mathtools}   
\usepackage{tablefootnote}
\usepackage{utfsym}

\newcommand{\cmark}{\ding{51}}
\newcommand{\xmark}{\ding{55}}

\usepackage{adjustbox}

\title{\textproc{HoH}: A Dynamic Benchmark for Evaluating the Impact of Outdated Information on Retrieval-Augmented Generation}

\author{%
  Jie Ouyang, Tingyue Pan, Mingyue Cheng\thanks{\ \ Corresponding author.}, Ruiran Yan, Yucong Luo, Jiaying Lin, Qi Liu
   \\
   State Key Lab of Cognitive Intelligence, University of Science and Technology of China \\ 
  \texttt{\{ouyang\_jie,pty12345,yanruiran,prime666,linjya\}@mail.ustc.edu.cn} \\
  \texttt{\{mycheng,qiliuql\}@ustc.edu.cn} \\
  }

\begin{document}
\maketitle
\begin{abstract}

While Retrieval-Augmented Generation (RAG) has emerged as an effective approach for addressing the knowledge outdating problem in Large Language Models (LLMs), it still faces a critical challenge: the prevalence of outdated information in knowledge bases. Current research primarily focuses on incorporating up-to-date information, yet the impact of outdated information coexisting in retrieval sources remains inadequately addressed. To bridge this gap, we introduce \textproc{HoH}, the first benchmark specifically designed to evaluate the impact of outdated information on RAG. Our benchmark leverages token-level \textit{diff} algorithms combined with LLM pipelines to efficiently create a large-scale QA dataset that accurately captures the evolution of temporal knowledge in real-world facts.
Through comprehensive experiments, we reveal that outdated information significantly degrades RAG performance in two critical ways: (1) it substantially reduces response accuracy by distracting models from correct information, and (2) it can mislead models into generating potentially harmful outputs, even when current information is available. Current RAG approaches struggle with both retrieval and generation aspects when handling outdated information. These findings highlight the urgent need for innovative solutions to address the temporal challenges in RAG.
Our code and data are available at \href{https://github.com/0russwest0/HoH}{https://github.com/0russwest0/HoH}.

\end{abstract}

\section{Introduction}

\begin{figure}
    \centering
    \vspace{-0.5cm}
    \includegraphics[width=1\linewidth]{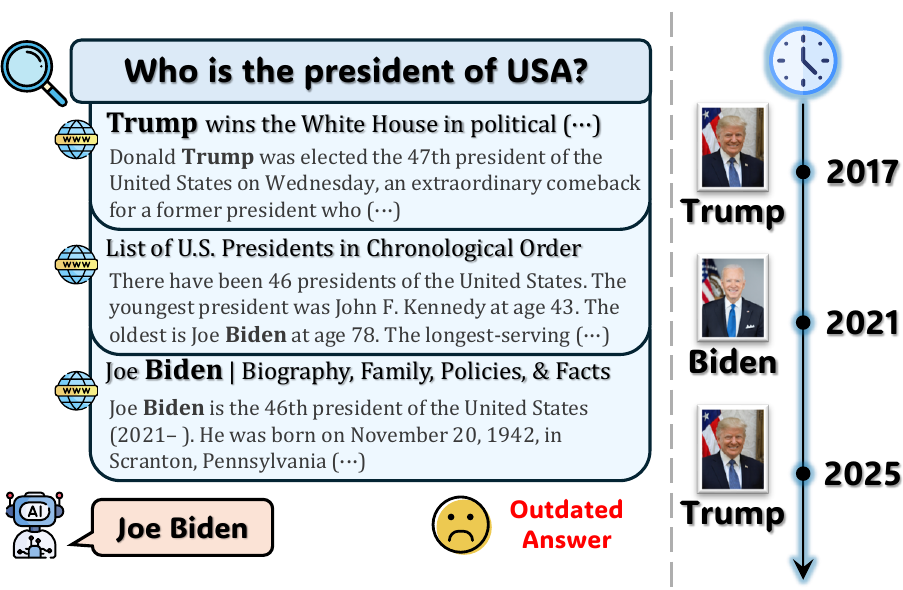}
    \caption{An illustration of how outdated information affects RAG. The example shows a query about the US president, where the retrieved passages contain outdated information, leading to potential confusion in generating an accurate response.}
    \label{fig:motivation}
    \vspace{-0.6cm}
\end{figure}

Large Language Models (LLMs) have revolutionized natural language processing with their remarkable knowledge retention and reasoning abilities \cite{openai2024gpt4technicalreport,grattafiori2024llama3herdmodels,qwen2.5}. However, the static nature of their pre-trained knowledge poses a significant challenge in handling rapidly evolving real-world information. LLMs frequently generate outdated responses that, while appearing plausible, no longer reflect current facts \cite{maynez2020faithfulness,liu2023evaluating}. Retrieval-Augmented Generation (RAG) \cite{lewis2020retrieval,guu2020retrieval,izacard-grave-2021-leveraging,borgeaud2022improving,yu2024knowledge} has emerged as a promising solution to this knowledge-outdating problem by dynamically retrieving and integrating up-to-date information from external sources.

Despite RAG's promise, a critical challenge remains unaddressed: the pervasive presence of outdated information in knowledge bases \cite{xin2024cost}. As illustrated in Figure 1, when querying about the current US president, RAG systems \cite{cheng2025survey,ouyang2024revisiting} may retrieve both current and outdated information, potentially leading to confusion or incorrect responses. This scenario is commonplace as information evolves and outdated facts inevitably accumulate across various sources, particularly in search engine scenarios where content is cached and redistributed. Current research primarily focuses on how to effectively retrieve and leverage the latest information \cite{karpukhin2020dense,chen2024bge}, while overlooking how the presence of outdated information could harm RAG. This oversight is significant, as even state-of-the-art industry RAG systems \cite{mehdi2023} frequently generate incorrect responses due to outdated information \cite{vu-etal-2024-freshllms}. To address this challenge, we introduce \textbf{\textproc{HoH}}
\footnote{\textproc{HoH} resembles a shocked emoji \includegraphics[scale=0.15]{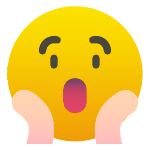}, expressing our alarm at the harm caused by outdated information.} 
(\textbf{H}ow \textbf{\textproc{o}}utdated information \textbf{H}arms Retrieval-Augmented Generation), the first large-scale benchmark designed to evaluate RAG's robustness against outdated information.

\textproc{HoH} comprises two primary components: \textproc{HoH-QA} and \textproc{HoH-SearchEngine}. \textproc{HoH-QA} is a continuously evolving QA dataset that maintains temporal alignment with contemporary world knowledge. Our dataset features comprehensive evidence annotations for each QA pair, enabling fine-grained evaluation of individual RAG components. More importantly, we introduce annotations for outdated answers and evidence for the first time, allowing systematic assessment of how outdated information impacts RAG performance. Complementing the QA dataset, \textproc{HoH-SearchEngine} simulates real-world search scenarios by maintaining both current and historical documents, better reflecting the actual challenges RAG faces in practice. Our benchmark currently includes 96,124 QA pairs and 219,463 documents, far surpassing the scale of existing datasets.

To ensure data quality and efficiency, we develop novel methods for dataset construction and maintenance. We propose a two-stage approach that combines \textit{diff} algorithms \cite{myers1986nd} with LLM pipelines to extract and verify factual changes from Wikipedia snapshots. This approach significantly outperforms previous methods \cite{jang-etal-2022-temporalwiki,kim2024carpe,ko-etal-2024-growover} in both efficiency and accuracy. The automated pipelines include rigorous quality control mechanisms, ensuring the reliability of the generated data while maintaining the dataset's continuous evolution.

Through extensive experiments on the \textproc{HoH} benchmark, we demonstrate that outdated information significantly degrades RAG performance in two critical ways: (1) outdated information substantially reduces response accuracy: even when current information is successfully retrieved, the mere presence of outdated information in the context leads to at least 20\% performance drop in mainstream LLMs, with some models performing worse than random guessing (-2.77\%); (2) outdated information can mislead models into generating harmful outputs: while models maintain appropriate uncertainty when no information is retrieved ("Unknown" or "Unsure"), they become highly prone to generating confident but incorrect responses when encountering outdated information. These findings highlight the urgent need for innovative solutions to address the temporal challenges in RAG.

In summary, our main contributions are:
\begin{enumerate}[itemsep=2pt, topsep=0pt, parsep=0pt]
    \item We introduce \textproc{HoH}, the first large-scale dynamic QA benchmark for evaluating RAG's robustness against outdated information, featuring comprehensive evidence annotations and a mock search engine that maintains both current and historical documents.
    
    \item We develop an efficient and scalable dataset construction method that combines traditional \textit{diff} algorithms with LLMs, achieving superior quality in factual change extraction while maintaining continuous dataset evolution.
    
    \item We present the first systematic analysis demonstrating the harmful impact of outdated information on RAG, providing crucial insights for future research in this domain.
\end{enumerate}

\begin{table*}[t]
\renewcommand{\arraystretch}{1.05}
\centering
\setlength{\tabcolsep}{3pt}
\caption{
    Comparison of our H\textproc{o}H with existing dynamic benchmarks. The automation indicates the feasibility of automatic generation. The Maintenance represents whether the validity of previously generated datasets is verified in the forthcoming time step. The evidence text indicates whether the dataset includes the evidence text. The outdated info shows whether the dataset contains annotation for outdate information. The 'question number' specifically counts Time-Sensitive Questions (excluding time-irrelevant ones), and 'article number' identifies the number of articles for retrieval, respectively.
    }
    \label{tab:comparison}
\begin{adjustbox}{width=\linewidth}
{\footnotesize
\begin{tabular}{@{}ccccccccc@{}}
\toprule
 & \textbf{Automation} & \textbf{Maintenance} & \textbf{Evidence Text} & \textbf{Outdated Info} & \textbf{Question Number} & \textbf{Article Number} \\ \midrule
\textbf{RealtimeQA} \cite{kasai2023realtime} & \xmark & \xmark & \xmark & \xmark & 2,340 & 16,023 \\
\textbf{FreshQA} \cite{vu-etal-2024-freshllms} & \xmark & \xmark & \xmark & \xmark & 600 & - \\
\textbf{TemporalWiki} \cite{jang-etal-2022-temporalwiki} & \cmark & \xmark & \xmark & \xmark & - & - \\
\textbf{EvolvingQA} \cite{kim2024carpe} & \cmark & \xmark & \xmark & \xmark & 23,283 & - \\
\textbf{GrowOVER} \cite{ko-etal-2024-growover} & \cmark & \cmark & \cmark & \xmark & 1,257 & - \\
\textbf{PAT-Questions} \cite{meem2024patquestionsselfupdatingbenchmarkpresentanchored} & \cmark & \cmark & \xmark & \xmark & 6,172 & - \\
\textbf{CLARK-New}s \cite{li2024languagemodelingeditableexternal} & \xmark & \xmark & \cmark & \cmark & 1,409 & 1,149 \\ \midrule
\textbf{H\textproc{o}H} (Ours) & \cmark & \cmark & \cmark & \cmark & 96,124 & 219,463 \\ \bottomrule
\end{tabular}}
\end{adjustbox}
\end{table*}

\section{Background and Related Work}

\textproc{H\textproc{o}H} belongs to a growing body of works aimed at handling Time-Sensitive Questions in Question Answering  (QA) systems. Time-Sensitive Questions are queries whose answers may vary depending on when the question is asked or the temporal context specified \cite{chen2021dataset}. Compared to traditional QA tasks \cite{dinanwizard,kwiatkowski2019natural}, Time-Sensitive QA demands the effective utilization of specific temporal contexts, encompassing multiple time-evolving facts, to address time-sensitive questions \cite{yang2024enhancing}. Based on whether temporal contexts are explicitly specified, Time-Sensitive Questions can be categorized into two types: Time-Situated Questions and Present-Anchored Questions.

\subsection{Time-Situated Question Answering}

Early research primarily focused on Time-Situated Questions \cite{chen2021dataset,zhang2021situatedqa,liska2022streamingqa,dhingra-etal-2022-time}, which contain explicit temporal contexts (e.g., "Who was the President of the USA in 2023?"). Subsequent works \cite{10.1145/3184558.3191536,saxena2021question,tan2023towards} introduced more complex temporal reasoning tasks. These datasets not only contain questions with temporal contexts but also require temporal reasoning capabilities (e.g., "Who was the President of the USA before Obama?"). While these benchmarks have advanced our understanding of temporal reasoning in QA systems, they do not address the challenges posed by continuously evolving information and the need for up-to-date answers in real-world applications.

\subsection{Present-Anchored Question Answering}

In contrast to Time-Situated Questions, Present-Anchored Questions implicitly require the most current information available (e.g., "Who is the President of the USA?"). These questions pose unique challenges as their answers may change over time, necessitating continuous updates to maintain accuracy. Recent years have witnessed several attempts to address these challenges through dynamic benchmarks. Early approaches, such as RealtimeQA and FreshQA \cite{kasai2023realtime,vu-etal-2024-freshllms}, rely on manual curation to ensure high-quality QA pairs, but suffer from limited scalability and impractical update frequencies. To overcome these limitations, automated approaches have emerged. TemporalWiki and EvolvingQA \cite{jang-etal-2022-temporalwiki,kim2024carpe} demonstrate the feasibility of automatic dataset generation, while GrowOVER and PAT-Questions \cite{ko-etal-2024-growover,meem2024patquestionsselfupdatingbenchmarkpresentanchored} advance this further by incorporating maintenance mechanisms. Recently, CLARK-News \cite{li2024languagemodelingeditableexternal}, in addition to providing the latest answers, is the first to record the historical progression of answer changes over time. However, due to its manual construction approach, it faces challenges in maintaining updates and is limited in scale.


As shown in Table \ref{tab:comparison}, our work, H\textproc{o}H, builds upon these foundations while addressing key limitations in existing benchmarks. Through automation and continuous maintenance, H\textproc{o}H ensures both scalability and data validity. It includes evidence text annotations, enabling fine-grained analyses of both retrieval and generation components in RAG systems. More importantly, it uniquely provides annotations for outdated information, enabling systematic study of its impact. Lastly, H\textproc{o}H offers unprecedented scale with 96,124 questions and 219,463 articles, surpassing existing benchmarks in comprehensiveness.
\begin{figure*}[h]
    \centering
    \includegraphics[width=1\linewidth]{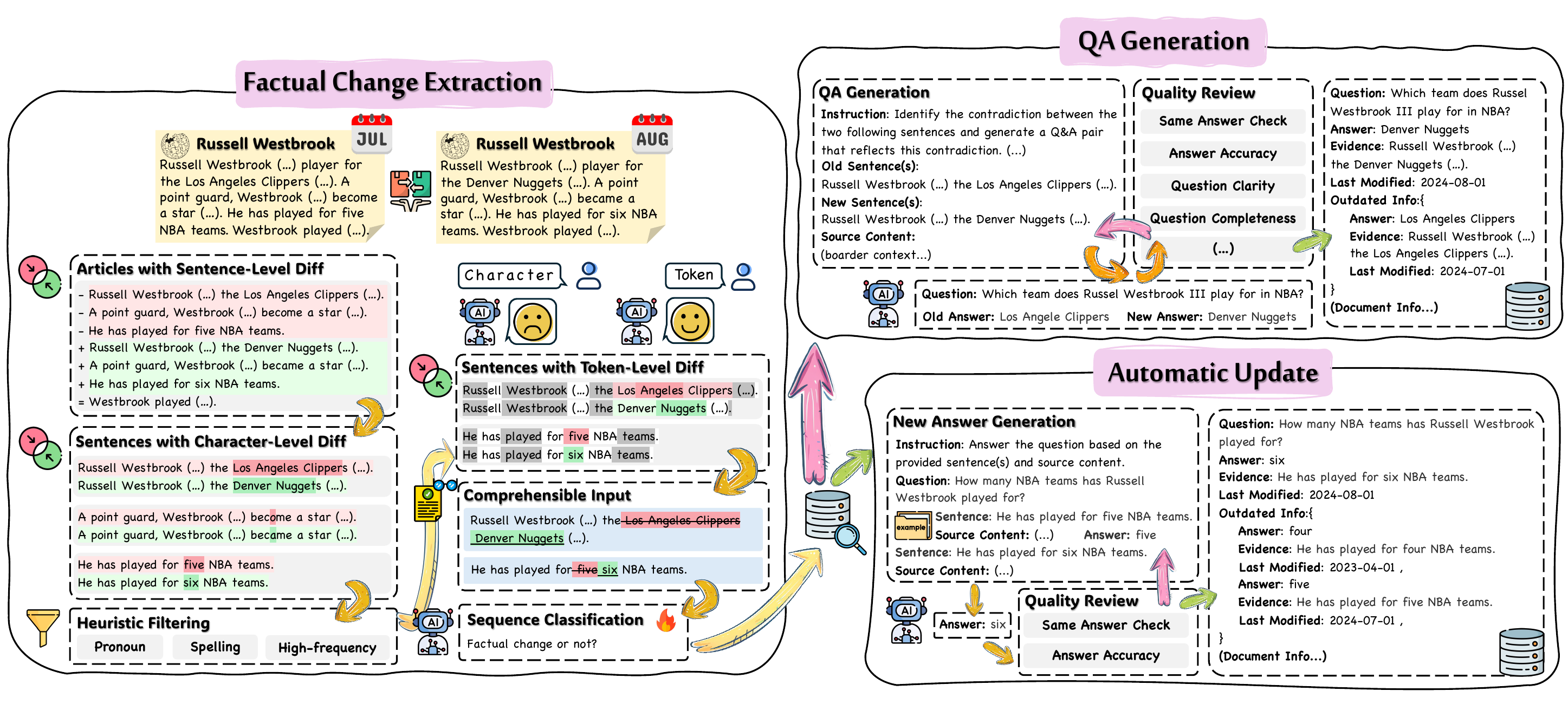}
    \caption{\small The overall process of \textproc{HoH-QA} construction comprises three primary parts: factual change extraction, QA generation, and automatic update. In the example shown, changes are identified between the July and August versions of a Wikipedia article. First, sentence pairs corresponding to factual changes are extracted. These extracted sentence pairs are then matched against previously generated datasets. For new sentence pairs, a QA generation process is conducted to create QA pairs. For sentence pairs that already exist, an automatic update process is employed to ensure the information remains current and accurate.} 
    \label{fig:overall process}
\end{figure*}

\section{The \textproc{HoH} Benchmark}


\textproc{HoH} benchmark comprises two key components: \textproc{HoH-QA}, a dynamic QA dataset that tracks temporal changes in facts, and \textproc{HoH-SearchEngine}, a mock search engine that simulates real-world information retrieval scenarios. In this section, we detail our methodology for constructing and maintaining these components.

\noindent \textbf{Article Selection}. \textproc{HoH} is based on Wikipedia snapshots \footnote{Download from \href{https://dumps.wikimedia.org/}{https://dumps.wikimedia.org/}}, which contain a vast amount of world knowledge. We select almost all articles from the snapshots, excluding only low-quality articles with excessively short content (less than 200 characters). The initial version was generated based on snapshots from 2024-06-01 and 2024-07-01, with regular monthly updates since then. As of now, we have updated it to the snapshot of 2024-11-01. For more detailed statistics, see Appendix \ref{app:dataset statistics}.

\subsection{\textproc{HoH-QA}}


\textproc{HoH-QA} is a dynamic open-domain QA dataset that tracks temporal changes in facts through two complementary mechanisms: (1) capturing time-sensitive questions whose ground-truth answers evolve over time, and (2) maintaining continuous dataset updates to reflect the latest world knowledge. Following \citet{jang-etal-2022-temporalwiki,kim2024carpe,ko-etal-2024-growover}, we generate \textproc{HoH-QA} through comparing two different Wikipedia snapshots at different time points. Building upon their approach, we enhance both the efficiency and quality, and additionally incorporate annotations for outdated information. 

As illustrated in Figure \ref{fig:overall process}, the main process comprises three primary parts: factual change extraction, QA generation, and automatic update. 

\noindent \textbf{Factual Change Extraction}. 
The initial phase of factual change extraction is to identify the modified sections across the same article of two different snapshots. We first split the old article $A_{old}$ and new article $A_{new}$ into sentences, get $A_{old}=\{s_{old_1},s_{old_2},\cdots,s_{old_n}\}$ and $A_{new}=\{s_{new_1},s_{new_2},\cdots,s_{new_m}\}$. Following \citet{kim2024carpe}, we employ the Myers \textit{Diff} \footnote{Implemented by \href{https://github.com/google/diff-match-patch}{diff-match-patch}} \cite{myers1986nd} algorithm to detect modified sentence pairs. \textit{Diff} algorithms are commonly employed to compare differences between texts, code (e.g., in Git), and are generally applicable to any sequence. By treating an article as a sequence of sentences, we obtain modified sentence pairs.

While prior works \cite{jang-etal-2022-temporalwiki,kim2024carpe} extract changes only at the sentence level, modified sentence pairs do not necessarily correspond to factual alterations.
We further filtered these pairs and applied a \textit{diff} algorithm at the character level to identify differences between sentences. For character-level differences, we employ heuristic methods to filter out modifications that are apparently not factual changes.

To further ensure the quality of extracted factual changes, we introduce a semantic screening stage using language models (LMs). Inspired by the human approach of reviewing different code versions, we consolidated the sentences before and after modification and delineated the differences between sentences in a comprehensible manner. For added parts, we use underlines for marking, and for deleted parts, we use strikethrough. To ensure that the marked differences align with the model's reading habits, we do not process differences at the character level. Instead, we combine the \textit{diff} algorithm with the tokenizer, handling differences at the token level.

In building a reliable screening model, we constructed a high-quality training dataset where three annotators independently verified 2,000 sentence pairs from our heuristically filtered samples. The resulting fine-tuned model\footnote{Based on Qwen2.5-0.5B \cite{qwen2.5}} achieves 96.8\% accuracy and 95.1\% F1 score in identifying genuine factual changes, outperforming previous methods by a significant margin. Complete details of the fine-tuning process and comparative evaluations are provided in Appendix \ref{app:fine-tuning} and \ref{app:experiment for SLM}.

\noindent \textbf{QA Generation}. 
Following established practices in automatic QA generation \cite{trischler2017newsqa,rajpurkar2018know}, we employ LLMs to create question-answer pairs from the extracted factual changes. For each identified change, we generate a question that captures the temporal aspect, along with two distinct answer versions: the current correct answer and its outdated counterpart. The corresponding sentence pairs serve as evidence, labeled as current and outdated evidence accordingly.


The quality of generated QA pairs is ensured through a multi-stage review process. After initial generation, LLMs conduct a thorough review, focusing on the quality of both questions and answers. Detailed generation prompts, quality control procedures, and a manual validation confirming the high reliability of this automated process are provided in Appendix \ref{app:details of qa generation}.

\noindent \textbf{Automatic Update}. Our dataset maintains synchronization with evolving knowledge through monthly updates based on new Wikipedia snapshots. For each update cycle, we apply our factual change extraction pipeline to identify modifications between consecutive snapshots. The update process handles two scenarios: (1) newly emerged facts, where we directly apply our QA generation pipeline, and (2) previously captured facts that undergo further changes, which require special handling.

For facts with continuous changes, we leverage LLMs to generate updated answers to existing questions using the latest information. This approach preserves the question's temporal relevance while creating a chain of answers that reflects the evolution of facts over time. Similar to our QA generation process, we employ LLMs to review the quality of updated answers. After confirming their accuracy and distinctness from previous versions, these answers are incorporated into our dataset, with each superseded answer preserved and marked as outdated.

\subsection{\textproc{HoH-SearchEngine}}


As QA datasets evolve over time, the corresponding external knowledge sources for retrieval must maintain temporal consistency. Search engines, being one of the most common forms of external knowledge sources, offer convenience for knowledge updates.
We implement \textproc{HoH-SearchEngine} using Elasticsearch\footnote{\href{https://www.elastic.co/elasticsearch}{https://www.elastic.co/elasticsearch}} \cite{elasticsearch2018elasticsearch}, extending its default BM25-based ranking \cite{robertson2009probabilistic} with temporal awareness. Specifically, we introduce a Gaussian decay function to discount outdated information, mimicking the temporal preference of real-world search engines. The underlying corpus comprises both current and historical versions of all articles used in the QA dataset construction, simulating the prevalence of outdated information on the web.
\section{Experiment Setup}

\begin{figure*}
    \centering
    \includegraphics[width=1\linewidth]{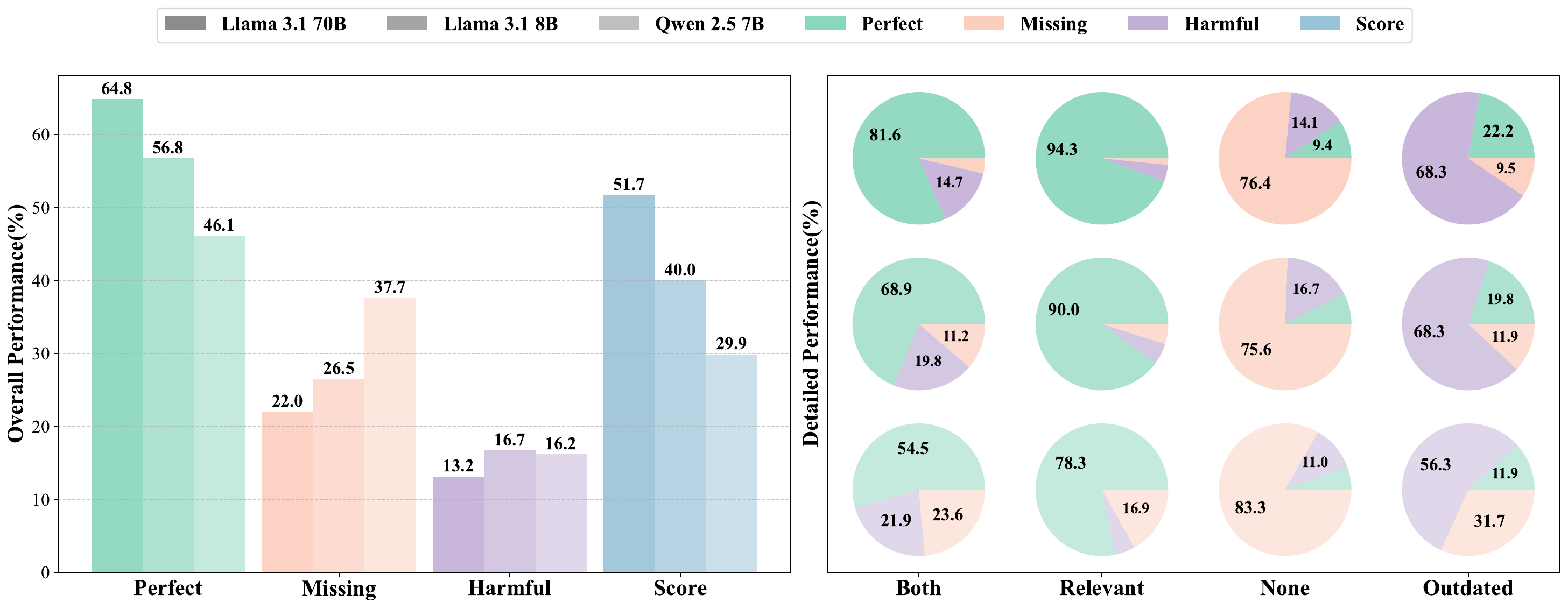}
    \caption{End-to-end performance comparison of different LLMs in the RAG evaluation. The left panel shows the overall performance metrics, while the right panel presents a detailed breakdown across categories.}
    \label{fig:rag}
\end{figure*}

In this section, we present definitions for document/passage categories, introduce time-aware instruction, explain the data and models tested in our experiments, and discuss the metrics and evaluation methods.

\subsection{Type of Documents/Passages}

In our study, we categorize documents/passages into three distinct types, each represented by a unique symbol, based on their relevance to the questions:

\textbf{Relevant ($\mathcal{R}$)}: These passages contain the correct answer and are useful for the question.

\textbf{Outdated ($\mathcal{O}$)}: These are previously relevant but have become obsolete, no longer accurate due to temporal changes.

\textbf{Distracting ($\mathcal{D}$)}: These documents are semantically similar to the question but do not contribute correct answers.


\subsection{Time-Aware Instruction and Setting}

Following \citet{vu-etal-2024-freshllms}, we have incorporated temporal information into the traditional RAG instruction, which is essential for addressing dynamic questions. We add the ``Current Date'' to the system prompt and included ``Last Modified Time'' for all documents, enabling temporal context awareness. Prior to response generation, the model is directed to prioritize the most current information. Figure \ref{fig:prompt for rag} shows the complete instruction example. Additionally, we've implemented a time-weighted search algorithm as the default mechanism for our \textproc{HoH-SearchEngine}.

\subsection{Data and Models}

\textbf{Evaluation Set}. We carefully curate a diverse subset of 10,000 QA samples by first clustering the entire dataset and then randomly selecting 1,000 samples from each of the 10 identified clusters.

\noindent \textbf{Model Tested}. We test several state-of-the-art embedding models and rerankers \cite{chen2024bge,xiao2024c}, including BGE-Base (\textit{bge-base-en-v1.5}), BGE-M3 (\textit{bge-m3}) and \textit{bge-reranker-v2-m3}. Additionally, we evaluate a range of cutting-edge LLMs \cite{qwen2.5,grattafiori2024llama3herdmodels} spanning from 7B to 70B parameters, such as Qwen-7B (\textit{Qwen-2.5-7B-Instruct}), Llama-8B (\textit{Llama-3.1-8B-Instruct}), and Llama-70B (\textit{Llama-3.1-70B-Instruct}).

\subsection{Metrics and Evaluation}

We evaluate responses using a model-based scoring system following \citet{yang2024crag}. Each answer is rated by LLMs as ``perfect'', ``missing'', or ``harmful'', scoring 1, 0, and -1, respectively. ``\textbf{Perfect}'' response accurately answers the question without hallucination. ``\textbf{Missing}'' indicates uncertainty, offering no clear benefit or detriment. ``\textbf{Harmful}'' denotes incorrect or misleading information. This method penalizes false content, prioritizing user safety by favoring omissions over errors.

\section{Experiments}

This section conducts a comprehensive benchmarking evaluation of various large language models and retrieval mechanisms within RAG. Through a series of experiments, our results reveal the limitations and shortcomings of existing methods when handling outdated information, providing insights for potential future enhancements.

\subsection{Overview of RAG Performance}

\textbf{Current RAG systems demonstrate poor overall performance, with even the most capable models achieving only mediocre results.} As shown on the left side of Figure \ref{fig:rag}, under our default setting, all three LLMs exhibit suboptimal performance. Even the most capable model, Llama-70B, achieves only a modest score of 51.7\%. The performance degradation is more pronounced in smaller models, with Llama-8B and Qwen-7B scoring merely 40.0\% and 29.9\%, respectively.

\textbf{The presence of outdated information significantly degrades model performance, causing not only reduced accuracy but also increased harmful outputs.} To analyze this phenomenon, we categorize the retrieved information based on whether it includes relevant ($\mathcal{R}$) and outdated ($\mathcal{O}$) passages: \textit{both} ($\mathcal{R}$ and $\mathcal{O}$), \textit{relevant} ($\mathcal{R}$ only), \textit{outdated} ($\mathcal{O}$ only) and \textit{none}. As illustrated in the right side of Figure \ref{fig:rag}, when only relevant information is retrieved, all models demonstrate relatively strong performance. However, the addition of outdated information substantially reduces accuracy and increases harmful outputs.

Most concerningly, outdated information ($\mathcal{O}$) poses a greater risk than having no relevant information at all. In scenarios where only $\mathcal{O}$ is retrieved, models exhibit dangerous overconfidence, frequently generating harmful misinformation. In contrast, when no $\mathcal{R}$ or $\mathcal{O}$ is retrieved, models show appropriate uncertainty, producing fewer harmful outputs despite low accuracy. This stark contrast suggests that outdated information actively misleads models rather than merely adding noise to the generation process.

\subsection{Analysis of Performance Degradation}

Our previous analysis reveals that outdated information poses a fundamental challenge to RAG, leading to not only degraded performance but also increased harmful outputs. To understand the root causes of this challenge, we conduct a systematic investigation from three perspectives: (1) the \textbf{retrieval} module's capability in distinguishing and filtering outdated information, (2) the \textbf{generation} module's robustness against outdated information, and (3) the underlying \textbf{time awareness} of large language models.

\begin{table}[t]
\caption{Hit rate of relevant ($\mathcal{R}$) and outdated ($\mathcal{O}$) documents for search engine results.}
\label{tab:search}
\small
\centering
\begin{tabular}{@{}cc|llll@{}}
\toprule
\textbf{Decay} & \textbf{Type} & \multicolumn{1}{c}{\textbf{@5}} & \multicolumn{1}{c}{\textbf{@10}} & \multicolumn{1}{c}{\textbf{@20}} & \multicolumn{1}{c}{\textbf{@50}} \\ \midrule
- & $\mathcal{R}$ & 0.8707 & 0.9131 & 0.9351 & 0.9551 \\
- & $\mathcal{O}$ & 0.8837 & 0.9159 & 0.9365 & 0.9747 \\
Gauss & $\mathcal{R}$ & 0.7023 & 0.7453 & 0.7842 & 0.8294 \\
Gauss & $\mathcal{O}$ & 0.4950 & 0.5457 & 0.5896 & 0.6463 \\ \bottomrule
\end{tabular}
\end{table}

\begin{table}[t]
\caption{Hit rate of relevant ($\mathcal{R}$) and outdated ($\mathcal{O}$) passages for embedding models and reranker.}
\label{tab:retrieval}
\centering
\small
\begin{tabular}{@{}cc|cc|c@{}}
\toprule
\multirow{2}{*}{\textbf{Embedding}} & \multirow{2}{*}{\textbf{Type}} & \multicolumn{2}{c|}{\textbf{w/o Rerank}} & \textbf{Rerank} \\ \cmidrule(l){3-5} 
 &  & \textbf{@5} & \textbf{@10} & \textbf{@5} \\ \midrule
\multirow{2}{*}{\textit{bge-base-en-v1.5}} & $\mathcal{R}$ & 0.7019 & 0.7365 & 0.7336 \\
 & $\mathcal{O}$ & 0.5289 & 0.5578 & 0.5563 \\
\multirow{2}{*}{\textit{bge-m3}} & $\mathcal{R}$ & 0.7312 & 0.7578 & 0.7754 \\
 & $\mathcal{O}$ & 0.5507 & 0.5698 & 0.5678 \\ \bottomrule
\end{tabular}
\end{table}

\begin{table*}[h]
\caption{Performance of the generation module with passages sorted by relevance score in descending order for $\{\mathcal{R} \times 1,\mathcal{D} \times n\}$. Per. (perfect), Mis. (missing), Har. (harmful) and Score are in percentage.}
\label{tab:generator_distracting}
\renewcommand{\arraystretch}{0.85}
\centering
\setlength{\tabcolsep}{6pt}
\scalebox{0.85}{
\begin{tabular}{@{}ccccccccccccc@{}}
\toprule
\multicolumn{1}{c|}{\textbf{}} & \multicolumn{4}{c|}{\textbf{Llama 3.1 70B}} & \multicolumn{4}{c|}{\textbf{Llama 3.1 8B}} & \multicolumn{4}{c}{\textbf{Qwen 2.5 7B}} \\ \midrule
\multicolumn{1}{c|}{\textbf{$n$}} & \textbf{Per.} & \textbf{Mis.} & \textbf{Har.} & \multicolumn{1}{c|}{\textbf{Score}} & \textbf{Per.} & \textbf{Mis.} & \textbf{Har.} & \multicolumn{1}{c|}{\textbf{Score}} & \textbf{Per.} & \textbf{Mis.} & \textbf{Har.} & \textbf{Score} \\ \midrule
\multicolumn{1}{c|}{\textbf{1}} & \textbf{93.24} & 2.64 & \textbf{4.12} & \multicolumn{1}{c|}{\textbf{89.12}} & \textbf{90.34} & \textbf{5.02} & \textbf{4.64} & \multicolumn{1}{c|}{\textbf{85.70}} & \textbf{79.82} & \textbf{16.67} & \textbf{3.51} & \textbf{76.31} \\
\multicolumn{1}{c|}{\textbf{2}} & 92.99 & 2.76 & 4.25 & \multicolumn{1}{c|}{88.74} & 89.43 & 5.63 & 4.94 & \multicolumn{1}{c|}{84.49} & 77.18 & 18.77 & 4.05 & 73.13 \\
\multicolumn{1}{c|}{\textbf{3}} & 92.73 & 2.71 & 4.56 & \multicolumn{1}{c|}{88.17} & 88.40 & 6.36 & 5.24 & \multicolumn{1}{c|}{83.16} & 76.21 & 19.74 & 4.05 & 72.16 \\
\multicolumn{1}{c|}{\textbf{4}} & 92.85 & \textbf{2.63} & 4.52 & \multicolumn{1}{c|}{88.33} & 87.87 & 6.89 & 5.24 & \multicolumn{1}{c|}{82.63} & 76.14 & 19.42 & 4.44 & 71.70 \\
\multicolumn{1}{c|}{\textbf{6}} & 92.57 & 2.84 & 4.59 & \multicolumn{1}{c|}{87.98} & 86.91 & 7.12 & 5.97 & \multicolumn{1}{c|}{80.94} & 74.86 & 20.52 & 4.62 & 70.24 \\\bottomrule
\end{tabular}
}
\end{table*}

\begin{table*}[h]
\caption{Performance of the generation module under different passage ordering strategies (by relevance score or date) for $\{\mathcal{R} \times 1,\mathcal{O} \times 1,\mathcal{D} \times n\}$. Score$\downarrow$ denotes sorting passages by relevance score in descending order, Date$\downarrow$ and Date$\uparrow$ denote sorting by date in descending and ascending order respectively. Per. (perfect), Mis. (missing), Har. (harmful) and Score are in percentage.}
\label{tab:generator_outdated}
\renewcommand{\arraystretch}{0.85}
\centering
\setlength{\tabcolsep}{6pt}
\scalebox{0.85}{
\begin{tabular}{@{}cccccccccccccc@{}}
\toprule
\textbf{} & \multicolumn{1}{c|}{\textbf{}} & \multicolumn{4}{c|}{\textbf{Llama 3.1 70B}} & \multicolumn{4}{c|}{\textbf{Llama 3.1 8B}} & \multicolumn{4}{c}{\textbf{Qwen 2.5 7B}} \\ \midrule
\multicolumn{1}{c|}{\textbf{Order}} & \multicolumn{1}{c|}{\textbf{$n$}} & \textbf{Per.} & \textbf{Mis.} & \textbf{Har.} & \multicolumn{1}{c|}{\textbf{Score}} & \textbf{Per.} & \textbf{Mis.} & \textbf{Har.} & \multicolumn{1}{c|}{\textbf{Score}} & \textbf{Per.} & \textbf{Mis.} & \textbf{Har.} & \textbf{Score} \\ \midrule
\multicolumn{1}{c|}{\multirow{5}{*}{\textbf{Score$\downarrow$}}} & \multicolumn{1}{c|}{\textbf{0}} & 83.00 & 6.95 & 10.05 & \multicolumn{1}{c|}{72.95} & 68.13 & 14.67 & 17.20 & \multicolumn{1}{c|}{50.93} & 55.61 & 22.71 & 21.68 & 33.93 \\
\multicolumn{1}{c|}{} & \multicolumn{1}{c|}{\textbf{1}} & 81.13 & 4.74 & 14.13 & \multicolumn{1}{c|}{67.00} & 66.87 & 14.44 & 18.69 & \multicolumn{1}{c|}{48.18} & 54.46 & 25.65 & 19.89 & 34.57 \\
\multicolumn{1}{c|}{} & \multicolumn{1}{c|}{\textbf{2}} & 80.83 & 4.56 & 14.61 & \multicolumn{1}{c|}{66.22} & 68.53 & 12.08 & 19.39 & \multicolumn{1}{c|}{49.14} & 53.73 & 26.37 & 19.90 & 33.83 \\
\multicolumn{1}{c|}{} & \multicolumn{1}{c|}{\textbf{3}} & 80.44 & 4.17 & 15.39 & \multicolumn{1}{c|}{65.05} & 68.45 & 11.55 & 20.00 & \multicolumn{1}{c|}{48.45} & 53.90 & 25.58 & 20.52 & 33.38 \\
\multicolumn{1}{c|}{} & \multicolumn{1}{c|}{\textbf{5}} & 79.77 & 4.34 & 15.89 & \multicolumn{1}{c|}{63.88} & 69.99 & 9.26 & 20.75 & \multicolumn{1}{c|}{49.24} & 54.29 & 24.86 & 20.85 & 33.44 \\ \midrule
\multicolumn{1}{c|}{\multirow{5}{*}{\textbf{Date$\downarrow$}}} & \multicolumn{1}{c|}{\textbf{0}} & \textbf{90.05} & 3.78 & \textbf{6.17} & \multicolumn{1}{c|}{\textbf{83.88}} & \textbf{71.32} & 12.91 & \textbf{15.77} & \multicolumn{1}{c|}{\textbf{55.55}} & 42.06 & 23.64 & 34.30 & 7.76 \\
\multicolumn{1}{c|}{} & \multicolumn{1}{c|}{\textbf{1}} & 88.39 & 2.67 & 8.94 & \multicolumn{1}{c|}{79.45} & 70.15 & 8.74 & 21.11 & \multicolumn{1}{c|}{49.04} & 41.42 & 23.85 & 34.73 & 6.69 \\
\multicolumn{1}{c|}{} & \multicolumn{1}{c|}{\textbf{2}} & 87.98 & \textbf{2.60} & 9.42 & \multicolumn{1}{c|}{78.56} & 68.65 & 7.08 & 24.27 & \multicolumn{1}{c|}{44.38} & 40.77 & 23.38 & 35.85 & 4.92 \\
\multicolumn{1}{c|}{} & \multicolumn{1}{c|}{\textbf{3}} & 87.12 & 2.70 & 10.18 & \multicolumn{1}{c|}{76.94} & 66.09 & 6.13 & 27.78 & \multicolumn{1}{c|}{38.31} & 39.77 & 21.79 & 38.44 & 1.33 \\
\multicolumn{1}{c|}{} & \multicolumn{1}{c|}{\textbf{5}} & 86.17 & 2.72 & 11.11 & \multicolumn{1}{c|}{75.06} & 63.97 & \textbf{4.52} & 31.51 & \multicolumn{1}{c|}{32.46} & 37.92 & \textbf{21.39} & 40.69 & -2.77 \\ \midrule
\multicolumn{1}{c|}{\multirow{5}{*}{\textbf{Date$\uparrow$}}} & \multicolumn{1}{c|}{\textbf{0}} & 75.07 & 10.90 & 14.03 & \multicolumn{1}{c|}{61.04} & 63.74 & 17.09 & 19.17 & \multicolumn{1}{c|}{44.57} & \textbf{70.04} & 21.46 & \textbf{8.50} & \textbf{61.54} \\
\multicolumn{1}{c|}{} & \multicolumn{1}{c|}{\textbf{1}} & 73.50 & 6.99 & 19.51 & \multicolumn{1}{c|}{53.99} & 62.01 & 16.46 & 21.53 & \multicolumn{1}{c|}{40.48} & 63.88 & 24.39 & 11.73 & 52.15 \\
\multicolumn{1}{c|}{} & \multicolumn{1}{c|}{\textbf{2}} & 74.44 & 5.51 & 20.05 & \multicolumn{1}{c|}{54.39} & 64.21 & 13.16 & 22.63 & \multicolumn{1}{c|}{41.58} & 61.71 & 24.87 & 13.42 & 48.29 \\
\multicolumn{1}{c|}{} & \multicolumn{1}{c|}{\textbf{3}} & 73.97 & 5.25 & 20.78 & \multicolumn{1}{c|}{53.19} & 64.79 & 12.66 & 22.55 & \multicolumn{1}{c|}{42.24} & 61.63 & 23.95 & 14.42 & 47.21 \\
\multicolumn{1}{c|}{} & \multicolumn{1}{c|}{\textbf{5}} & 73.23 & 4.72 & 22.05 & \multicolumn{1}{c|}{51.18} & 64.95 & 9.97 & 25.08 & \multicolumn{1}{c|}{39.87} & 60.40 & 23.20 & 16.40 & 44.00 \\ \bottomrule
\end{tabular}
}
\end{table*}

\textbf{Retrieval}. We first investigate the ability of the retrieval module to distinguish outdated information. Ideally, we expect relevant information ($\mathcal{R}$) to rank as high as possible while outdated information ($\mathcal{O}$) should rank lower.

\textbf{Traditional search engines struggle to effectively filter out outdated information while maintaining high recall of relevant content.} Our empirical investigation of search engine performance compares results with and without temporal relevance considerations (Table \ref{tab:search}). When temporal relevance is not considered, outdated information ($\mathcal{O}$) actually shows slightly higher retrieval rates than relevant information ($\mathcal{R}$). After incorporating temporal relevance, while $\mathcal{R}$'s ranking improves relative to $\mathcal{O}$, there remains a nearly 50\% probability of retrieving outdated information in the top 5 results. Moreover, this temporal awareness comes at the cost of significantly reduced recall for relevant information, with $\mathcal{R}$'s hit rate dropping by approximately 17\% in top 5 results.

\textbf{Current embedding models and rerankers, while effective at capturing semantic relevance, show limited ability in distinguishing temporal validity.} As shown in Table \ref{tab:retrieval}, all tested models demonstrate strong performance in retrieving relevant information, with BGE-M3 outperforming BGE-Base, and both models showing improvements after reranking. However, these models simultaneously maintain high retrieval rates for outdated information, with hit rates consistently exceeding 50\%. More problematically, models that perform better at retrieving relevant information also show proportionally better performance at retrieving outdated information, suggesting a fundamental limitation in their ability to distinguish temporal relevance.

\textbf{These findings indicate an inherent conflict between maximizing relevant information retrieval and minimizing outdated content retrieval.} Current mainstream methodologies appear to treat temporal relevance as a traditional relevance problem, leading to a forced trade-off between these competing objectives. This limitation at the retrieval stage places a heavy burden on the subsequent generation module to correctly identify and handle outdated information.

\begin{table*}[h]
\caption{Analysis of the Correlation (\%) between LLMs' Timeliness Awareness and RAG Performance.}
\label{tab:timelineness_awareness}
\renewcommand{\arraystretch}{0.85}
\centering
\setlength{\tabcolsep}{6pt}
\scalebox{0.85}{
\begin{tabular}{@{}cc|ccc|ccc|ccc@{}}
\toprule
\multicolumn{2}{c|}{\textbf{}} & \multicolumn{3}{c|}{\textbf{Llama 3.1 70B}} & \multicolumn{3}{c|}{\textbf{Llama 3.1 8B}} & \multicolumn{3}{c}{\textbf{Qwen 2.5 7B}} \\ \midrule
$\mathcal{A_C}$ & $\mathcal{A_O}$ & \textbf{Perfect} & \textbf{Missing} & \textbf{Harmful} & \textbf{Perfect} & \textbf{Missing} & \textbf{Harmful} & \textbf{Perfect} & \textbf{Missing} & \textbf{Harmful} \\ \midrule
\xmark & \xmark & 24.70 & 2.63 & 72.67 & 33.81 & 15.75 & 50.44 & 26.67 & 23.92 & 49.41 \\
\xmark & \cmark & 52.42 & 11.64 & 35.94 & 52.61 & 23.12 & 24.26 & 44.30 & 33.95 & 21.75 \\
\cmark & \xmark & 76.49 & \textbf{0.25} & 23.25 & 68.77 & \textbf{7.59} & 23.64 & 61.25 & \textbf{16.45} & 22.30 \\
\cmark & \cmark & \textbf{93.24} & 2.54 & \textbf{4.22} & \textbf{83.86} & 8.67 & \textbf{7.47} & \textbf{76.75} & 16.88 & \textbf{6.37} \\ \bottomrule
\end{tabular}
}
\end{table*}

\textbf{Generation}. Given the retrieval phase's limitations in managing outdated information, we explore retaining such data and relying on robust generative models for its identification and processing.

\textbf{The generation module is highly sensitive to outdated information, showing far greater degradation compared to distracting passages.} As shown in Table \ref{tab:generator_distracting}, while increasing the number of distracting passages ($\mathcal{D}$) does lead to performance decline, the impact is relatively modest. For Llama-70B, even with six distracting passages, accuracy and harmful output rates vary by less than 1\%, with overall scores decreasing by under 2\%. However, when introducing just one outdated passage (Table \ref{tab:generator_outdated}:Score$\downarrow$), the impact is profound – Llama-70B's perfect scores drop by over 10\%, harmful outputs increase by up to 11\%, and overall scores decrease by more than 24\%.

\textbf{Less capable models demonstrate even greater vulnerability to outdated information.} The smaller models – Llama-8B and Qwen-7B – experience severe performance degradation with outdated information present. Their perfect scores fall by approximately 20\%, and harmful outputs increase by up to 18\%. This heightened sensitivity suggests that model capacity plays a crucial role in handling outdated information, though even the most capable models struggle significantly.

\textbf{The ordering of passages critically influences model performance, with different models showing varying levels of sensitivity to passage arrangement.} Table \ref{tab:generator_outdated} reveals diverse responses to different ordering strategies across models. Llama-8B shows moderate robustness with score fluctuations within 11\%, while Llama-70B maintains better overall performance but experiences variations up to 25\%. Most concerningly, Qwen-7B displays extreme sensitivity with score variations exceeding 50\%, sometimes performing worse than random guessing (-2.77\%). These findings suggest that optimizing passage ordering could be crucial for improving RAG performance, though it cannot fully mitigate the fundamental challenge posed by outdated information.

\textbf{Timeliness Awareness}. The limitations observed in both retrieval and generation modules point to a more fundamental question: to what extent can LLMs understand and reason about temporal information? This investigation is crucial as it may explain the root cause of RAG's vulnerability to outdated information. To answer this question, we conduct a systematic analysis of LLMs' timeliness awareness capabilities.

We decompose timeliness awareness into two components: Current Awareness ($\mathcal{A_C}$), which measures the model's ability to recognize current information as up-to-date, and Outdated Awareness ($\mathcal{A_O}$), which evaluates the model's capability to identify outdated information as not current. These two types of awareness form the foundation for analyzing temporal understanding in LLMs.

\begin{figure}[t]
    \centering
    \includegraphics[width=1\linewidth]{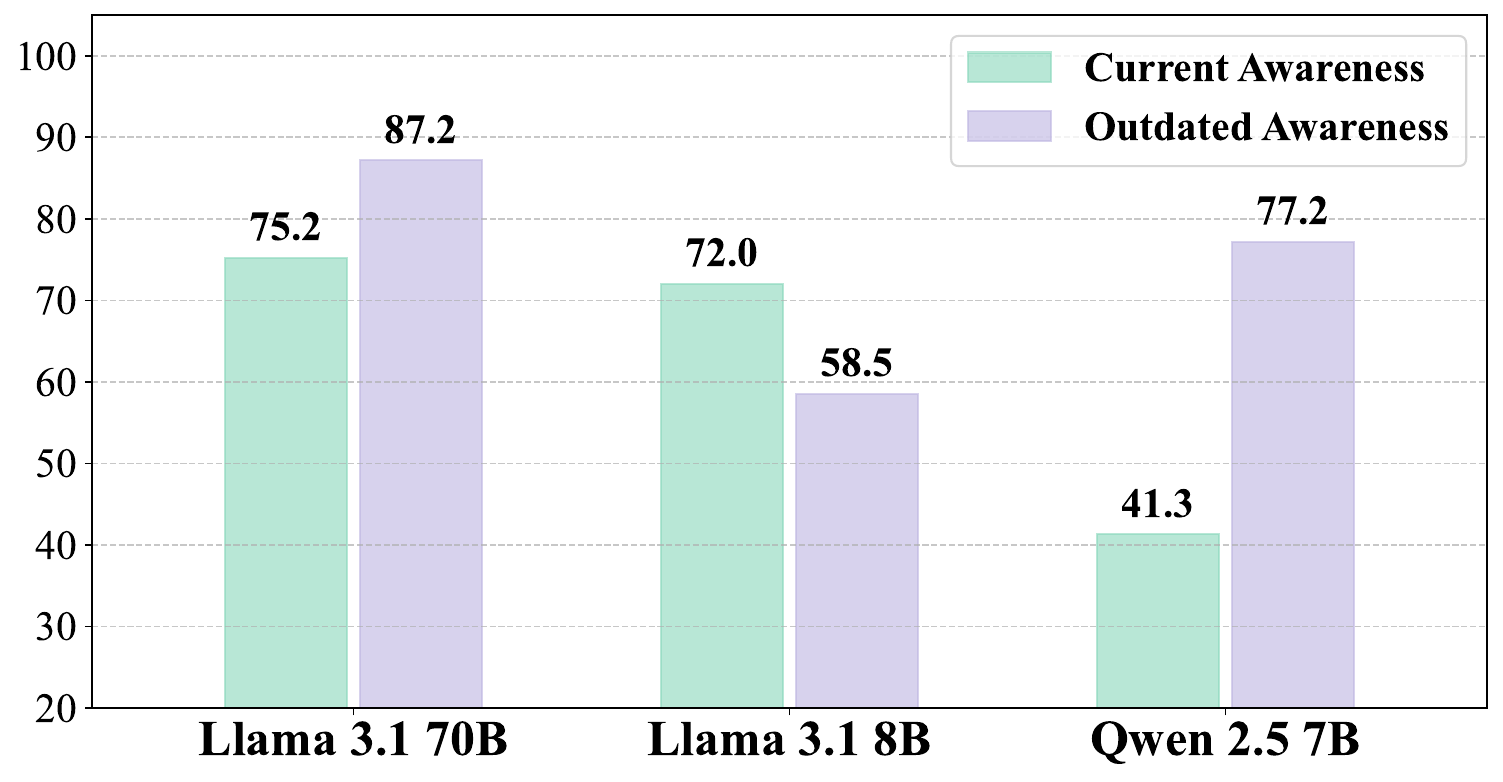}
    \caption{Timeliness awareness (\%) of LLMs ideally, where retrieval results always contain both $\mathcal{R}$ and $\mathcal{O}$.}
    \label{fig:timeliness}
\end{figure}

\textbf{Current LLMs demonstrate inadequate timeliness awareness, struggling to consistently differentiate between current and outdated information.} As shown in Figure \ref{fig:timeliness}, even in an ideal scenario where models have access to both relevant and outdated information, all tested models perform poorly in timeliness awareness. While Llama-70B shows relatively balanced performance in both aspects, smaller models demonstrate clear weaknesses - Llama-8B particularly struggles with $\mathcal{A_O}$ (58.5\%), while Qwen-7B shows poor $\mathcal{A_O}$ (41.3\%).

\textbf{Timeliness awareness capabilities strongly correlate with the performance of RAG.} As shown in Table \ref{tab:timelineness_awareness}, models exhibit a clear performance hierarchy based on their timeliness awareness. Models possessing both types of awareness consistently achieve the best performance, while those with single awareness show moderate degradation. Most concerningly, models lacking both types of awareness perform the worst, with dramatic increases in harmful outputs. This consistent pattern suggests that enhancing LLMs' inherent timeliness awareness could be crucial for improving the performance of RAG.

\textbf{An intriguing gap exists between outdated awareness and harmful output prevention.} Notably, some models with only $\mathcal{A_O}$ produce more harmful outputs than those with only $\mathcal{A_C}$, despite being able to recognize outdated information. Llama-70B demonstrates this paradox strikingly, generating 11\% more harmful outputs with $\mathcal{A_O}$ compared to $\mathcal{A_C}$. This suggests that merely identifying information as outdated does not guarantee the model will avoid using it harmfully. Further alignment might be needed to ensure models refrain from generating harmful outputs when they recognize the information as outdated.

\section{Conclusion}


We introduced \textproc{HoH}, the first benchmark designed to evaluate how outdated information impacts RAG systems. By novelly combining token-level \textit{diff} algorithms and language models, we efficiently constructed a high-quality large-scale QA dataset. Our experiments demonstrated that outdated information significantly degrades RAG performance by reducing accuracy and potentially generating harmful outputs. Current RAG approaches struggle with both retrieval and generation aspects when handling outdated information. This work provides a new perspective on RAG vulnerabilities and offers crucial infrastructure for future improvements.


\section{Limitation}

We use changes in Wikipedia snapshots to generate and update benchmarks, which effectively capture the evolution of real-world knowledge but also have several limitations. Firstly, Wikipedia snapshots have a fixed update frequency, while real-world knowledge changes continuously at varying speeds. For rapidly evolving fields like the stock market, our benchmarks cannot reflect real-time changes. Secondly, our QA datasets are primarily derived from individual articles, which may limit their effectiveness for tasks that require synthesizing information from multiple sources. Lastly, despite our efforts to closely simulate a real-world search engine, our mock search engine has significant limitations. This is mainly due to the homogenization of outdated information, as our outdated corpora primarily come from the historical versions of the same article, making them very similar to each other. In contrast, real-world outdated information is diverse, often sourced from various origins, and displays significant differences.

\section*{Acknowledgments}
This research was supported by grants from the National Key Research and Development Program of China (Grant No. 2024YFC3308200), the grants of the Provincial Natural Science Foundation of Anhui Province (No.2408085QF193), the Key Technologies R \& D Program of Anhui Province (No. 202423k09020039) and the Fundamental Research Funds for the Central Universities (No. WK2150110032).

\clearpage

\bibliography{main}

\begin{thebibliography}{46}
\providecommand{\natexlab}[1]{#1}

\bibitem[{Borgeaud et~al.(2022)Borgeaud, Mensch, Hoffmann, Cai, Rutherford, Millican, Van Den~Driessche, Lespiau, Damoc, Clark et~al.}]{borgeaud2022improving}
Sebastian Borgeaud, Arthur Mensch, Jordan Hoffmann, Trevor Cai, Eliza Rutherford, Katie Millican, George~Bm Van Den~Driessche, Jean-Baptiste Lespiau, Bogdan Damoc, Aidan Clark, et~al. 2022.
\newblock Improving language models by retrieving from trillions of tokens.
\newblock In \emph{International conference on machine learning}, pages 2206--2240. PMLR.

\bibitem[{Chen et~al.(2024)Chen, Xiao, Zhang, Luo, Lian, and Liu}]{chen2024bge}
Jianlv Chen, Shitao Xiao, Peitian Zhang, Kun Luo, Defu Lian, and Zheng Liu. 2024.
\newblock Bge m3-embedding: Multi-lingual, multi-functionality, multi-granularity text embeddings through self-knowledge distillation.
\newblock \emph{arXiv preprint arXiv:2402.03216}.

\bibitem[{Chen et~al.(2021)Chen, Wang, and Wang}]{chen2021dataset}
Wenhu Chen, Xinyi Wang, and William~Yang Wang. 2021.
\newblock A dataset for answering time-sensitive questions.
\newblock \emph{arXiv preprint arXiv:2108.06314}.

\bibitem[{Cheng et~al.(2025)Cheng, Luo, Ouyang, Liu, Liu, Li, Yu, Zhang, Cao, Ma et~al.}]{cheng2025survey}
Mingyue Cheng, Yucong Luo, Jie Ouyang, Qi~Liu, Huijie Liu, Li~Li, Shuo Yu, Bohou Zhang, Jiawei Cao, Jie Ma, et~al. 2025.
\newblock A survey on knowledge-oriented retrieval-augmented generation.
\newblock \emph{arXiv preprint arXiv:2503.10677}.

\bibitem[{Dhingra et~al.(2022)Dhingra, Cole, Eisenschlos, Gillick, Eisenstein, and Cohen}]{dhingra-etal-2022-time}
Bhuwan Dhingra, Jeremy~R. Cole, Julian~Martin Eisenschlos, Daniel Gillick, Jacob Eisenstein, and William~W. Cohen. 2022.
\newblock \href {https://doi.org/10.1162/tacl_a_00459} {Time-aware language models as temporal knowledge bases}.
\newblock \emph{Transactions of the Association for Computational Linguistics}, 10:257--273.

\bibitem[{Dinan et~al.(2018)Dinan, Roller, Shuster, Fan, Auli, and Weston}]{dinanwizard}
Emily Dinan, Stephen Roller, Kurt Shuster, Angela Fan, Michael Auli, and Jason Weston. 2018.
\newblock Wizard of wikipedia: Knowledge-powered conversational agents.
\newblock In \emph{International Conference on Learning Representations}.

\bibitem[{Elasticsearch(2018)}]{elasticsearch2018elasticsearch}
BV~Elasticsearch. 2018.
\newblock Elasticsearch.
\newblock \emph{software], version}, 6(1).

\bibitem[{Gade and Jetcheva(2024)}]{gade2024itstimeincorporatingtemporality}
Anoushka Gade and Jorjeta Jetcheva. 2024.
\newblock \href {https://arxiv.org/abs/2401.13222} {It's about time: Incorporating temporality in retrieval augmented language models}.
\newblock \emph{Preprint}, arXiv:2401.13222.

\bibitem[{Gao et~al.(2021)Gao, Yao, and Chen}]{gao2021simcse}
T~Gao, X~Yao, and Danqi Chen. 2021.
\newblock Simcse: Simple contrastive learning of sentence embeddings.
\newblock In \emph{EMNLP 2021-2021 Conference on Empirical Methods in Natural Language Processing, Proceedings}.

\bibitem[{Grattafiori et~al.(2024)Grattafiori, Dubey, and et~al.}]{grattafiori2024llama3herdmodels}
Aaron Grattafiori, Abhimanyu Dubey, and Abhinav~Jauhri et~al. 2024.
\newblock \href {https://arxiv.org/abs/2407.21783} {The llama 3 herd of models}.
\newblock \emph{Preprint}, arXiv:2407.21783.

\bibitem[{Guu et~al.(2020)Guu, Lee, Tung, Pasupat, and Chang}]{guu2020retrieval}
Kelvin Guu, Kenton Lee, Zora Tung, Panupong Pasupat, and Mingwei Chang. 2020.
\newblock Retrieval augmented language model pre-training.
\newblock In \emph{International conference on machine learning}, pages 3929--3938. PMLR.

\bibitem[{Izacard and Grave(2021)}]{izacard-grave-2021-leveraging}
Gautier Izacard and Edouard Grave. 2021.
\newblock \href {https://doi.org/10.18653/v1/2021.eacl-main.74} {Leveraging passage retrieval with generative models for open domain question answering}.
\newblock In \emph{Proceedings of the 16th Conference of the European Chapter of the Association for Computational Linguistics: Main Volume}, pages 874--880, Online. Association for Computational Linguistics.

\bibitem[{Jang et~al.(2022)Jang, Ye, Lee, Yang, Shin, Han, Kim, and Seo}]{jang-etal-2022-temporalwiki}
Joel Jang, Seonghyeon Ye, Changho Lee, Sohee Yang, Joongbo Shin, Janghoon Han, Gyeonghun Kim, and Minjoon Seo. 2022.
\newblock \href {https://doi.org/10.18653/v1/2022.emnlp-main.418} {{T}emporal{W}iki: A lifelong benchmark for training and evaluating ever-evolving language models}.
\newblock In \emph{Proceedings of the 2022 Conference on Empirical Methods in Natural Language Processing}, pages 6237--6250, Abu Dhabi, United Arab Emirates. Association for Computational Linguistics.

\bibitem[{Jia et~al.(2018)Jia, Abujabal, Saha~Roy, Str\"{o}tgen, and Weikum}]{10.1145/3184558.3191536}
Zhen Jia, Abdalghani Abujabal, Rishiraj Saha~Roy, Jannik Str\"{o}tgen, and Gerhard Weikum. 2018.
\newblock \href {https://doi.org/10.1145/3184558.3191536} {Tempquestions: A benchmark for temporal question answering}.
\newblock In \emph{Companion Proceedings of the The Web Conference 2018}, WWW '18, page 1057–1062, Republic and Canton of Geneva, CHE. International World Wide Web Conferences Steering Committee.

\bibitem[{Karpukhin et~al.(2020)Karpukhin, Oguz, Min, Lewis, Wu, Edunov, Chen, and Yih}]{karpukhin2020dense}
Vladimir Karpukhin, Barlas Oguz, Sewon Min, Patrick Lewis, Ledell Wu, Sergey Edunov, Danqi Chen, and Wen-tau Yih. 2020.
\newblock Dense passage retrieval for open-domain question answering.
\newblock In \emph{Proceedings of the 2020 Conference on Empirical Methods in Natural Language Processing (EMNLP)}, pages 6769--6781.

\bibitem[{Kasai et~al.(2023)Kasai, Sakaguchi, Takahashi, Le~Bras, Asai, Yu, Radev, Smith, Choi, and Inui}]{kasai2023realtime}
Jungo Kasai, Keisuke Sakaguchi, Yoichi Takahashi, Ronan Le~Bras, Akari Asai, Xinyan~Velocity Yu, Dragomir Radev, Noah~A Smith, Yejin Choi, and Kentaro Inui. 2023.
\newblock Realtime qa: what's the answer right now?
\newblock In \emph{Proceedings of the 37th International Conference on Neural Information Processing Systems}, pages 49025--49043.

\bibitem[{Kim et~al.(2024)Kim, Yoon, Ye, Bae, Ho, Hwang, and Yun}]{kim2024carpe}
Yujin Kim, Jaehong Yoon, Seonghyeon Ye, Sangmin Bae, Namgyu Ho, Sung~Ju Hwang, and Se-Young Yun. 2024.
\newblock Carpe diem: On the evaluation of world knowledge in lifelong language models.
\newblock In \emph{Proceedings of the 2024 Conference of the North American Chapter of the Association for Computational Linguistics: Human Language Technologies (Volume 1: Long Papers)}, pages 5401--5415.

\bibitem[{Ko et~al.(2024)Ko, Kim, Choi, and Kim}]{ko-etal-2024-growover}
Dayoon Ko, Jinyoung Kim, Hahyeon Choi, and Gunhee Kim. 2024.
\newblock \href {https://doi.org/10.18653/v1/2024.acl-long.181} {{G}row{OVER}: How can {LLM}s adapt to growing real-world knowledge?}
\newblock In \emph{Proceedings of the 62nd Annual Meeting of the Association for Computational Linguistics (Volume 1: Long Papers)}, pages 3282--3308, Bangkok, Thailand. Association for Computational Linguistics.

\bibitem[{Kwiatkowski et~al.(2019)Kwiatkowski, Palomaki, Redfield, Collins, Parikh, Alberti, Epstein, Polosukhin, Devlin, Lee et~al.}]{kwiatkowski2019natural}
Tom Kwiatkowski, Jennimaria Palomaki, Olivia Redfield, Michael Collins, Ankur Parikh, Chris Alberti, Danielle Epstein, Illia Polosukhin, Jacob Devlin, Kenton Lee, et~al. 2019.
\newblock Natural questions: A benchmark for question answering research.
\newblock \emph{Transactions of the Association for Computational Linguistics}, 7:452--466.

\bibitem[{Lazaridou et~al.(2022)Lazaridou, Gribovskaya, Stokowiec, and Grigorev}]{lazaridou2022internet}
Angeliki Lazaridou, Elena Gribovskaya, Wojciech Stokowiec, and Nikolai Grigorev. 2022.
\newblock Internet-augmented language models through few-shot prompting for open-domain question answering.
\newblock \emph{arXiv preprint arXiv:2203.05115}.

\bibitem[{Lewis et~al.(2020)Lewis, Perez, Piktus, Petroni, Karpukhin, Goyal, K{\"u}ttler, Lewis, Yih, Rockt{\"a}schel et~al.}]{lewis2020retrieval}
Patrick Lewis, Ethan Perez, Aleksandra Piktus, Fabio Petroni, Vladimir Karpukhin, Naman Goyal, Heinrich K{\"u}ttler, Mike Lewis, Wen-tau Yih, Tim Rockt{\"a}schel, et~al. 2020.
\newblock Retrieval-augmented generation for knowledge-intensive nlp tasks.
\newblock \emph{Advances in Neural Information Processing Systems}, 33:9459--9474.

\bibitem[{Li et~al.(2024)Li, Liu, Ross, Zeitoun, Neubig, and Andreas}]{li2024languagemodelingeditableexternal}
Belinda~Z. Li, Emmy Liu, Alexis Ross, Abbas Zeitoun, Graham Neubig, and Jacob Andreas. 2024.
\newblock \href {https://arxiv.org/abs/2406.11830} {Language modeling with editable external knowledge}.
\newblock \emph{Preprint}, arXiv:2406.11830.

\bibitem[{Liska et~al.(2022)Liska, Kocisky, Gribovskaya, Terzi, Sezener, Agrawal, Cyprien De~Masson, Scholtes, Zaheer, Young et~al.}]{liska2022streamingqa}
Adam Liska, Tomas Kocisky, Elena Gribovskaya, Tayfun Terzi, Eren Sezener, Devang Agrawal, D’Autume Cyprien De~Masson, Tim Scholtes, Manzil Zaheer, Susannah Young, et~al. 2022.
\newblock Streamingqa: A benchmark for adaptation to new knowledge over time in question answering models.
\newblock In \emph{International Conference on Machine Learning}, pages 13604--13622. PMLR.

\bibitem[{Liu et~al.(2023)Liu, Zhang, and Liang}]{liu2023evaluating}
Nelson~F Liu, Tianyi Zhang, and Percy Liang. 2023.
\newblock Evaluating verifiability in generative search engines.
\newblock In \emph{Findings of the Association for Computational Linguistics: EMNLP 2023}, pages 7001--7025.

\bibitem[{Maynez et~al.(2020)Maynez, Narayan, Bohnet, and McDonald}]{maynez2020faithfulness}
Joshua Maynez, Shashi Narayan, Bernd Bohnet, and Ryan McDonald. 2020.
\newblock On faithfulness and factuality in abstractive summarization.
\newblock In \emph{Proceedings of the 58th Annual Meeting of the Association for Computational Linguistics}, pages 1906--1919.

\bibitem[{Meem et~al.(2024)Meem, Rashid, Dong, and Hristidis}]{meem2024patquestionsselfupdatingbenchmarkpresentanchored}
Jannat~Ara Meem, Muhammad~Shihab Rashid, Yue Dong, and Vagelis Hristidis. 2024.
\newblock \href {https://arxiv.org/abs/2402.11034} {Pat-questions: A self-updating benchmark for present-anchored temporal question-answering}.
\newblock \emph{Preprint}, arXiv:2402.11034.

\bibitem[{Mehdi(2023)}]{mehdi2023}
Yusuf Mehdi. 2023.
\newblock \href {https://blogs.bing.com/search/march_2023/The-New-Bing-and-Edge-\%E2\%80\%93-Momentum-from-Our-First-Month} {The new {B}ing and {E}dge – progress from our first month | {B}ing search blog}.
\newblock Accessed on March 28, 2023.

\bibitem[{Myers(1986)}]{myers1986nd}
Eugene~W Myers. 1986.
\newblock An o (nd) difference algorithm and its variations.
\newblock \emph{Algorithmica}, 1(1):251--266.

\bibitem[{OpenAI et~al.(2024)OpenAI, Achiam, Adler, and et~al.}]{openai2024gpt4technicalreport}
OpenAI, Josh Achiam, Steven Adler, and Sandhini~Agarwal et~al. 2024.
\newblock \href {https://arxiv.org/abs/2303.08774} {Gpt-4 technical report}.
\newblock \emph{Preprint}, arXiv:2303.08774.

\bibitem[{Ouyang et~al.(2024)Ouyang, Luo, Cheng, Wang, Yu, Liu, and Chen}]{ouyang2024revisiting}
Jie Ouyang, Yucong Luo, Mingyue Cheng, Daoyu Wang, Shuo Yu, Qi~Liu, and Enhong Chen. 2024.
\newblock Revisiting the solution of meta kdd cup 2024: Crag.
\newblock \emph{arXiv preprint arXiv:2409.15337}.

\bibitem[{Paszke et~al.(2019)Paszke, Gross, Massa, Lerer, Bradbury, Chanan, Killeen, Lin, Gimelshein, Antiga et~al.}]{paszke2019pytorch}
Adam Paszke, Sam Gross, Francisco Massa, Adam Lerer, James Bradbury, Gregory Chanan, Trevor Killeen, Zeming Lin, Natalia Gimelshein, Luca Antiga, et~al. 2019.
\newblock Pytorch: An imperative style, high-performance deep learning library.
\newblock \emph{Advances in neural information processing systems}, 32.

\bibitem[{Qwen(2024)}]{qwen2.5}
Team Qwen. 2024.
\newblock \href {https://qwenlm.github.io/blog/qwen2.5/} {Qwen2.5: A party of foundation models}.

\bibitem[{Rajpurkar et~al.(2018)Rajpurkar, Jia, and Liang}]{rajpurkar2018know}
Pranav Rajpurkar, Robin Jia, and Percy Liang. 2018.
\newblock Know what you don’t know: Unanswerable questions for squad.
\newblock In \emph{Proceedings of the 56th Annual Meeting of the Association for Computational Linguistics (Volume 2: Short Papers)}, pages 784--789.

\bibitem[{Robertson et~al.(2009)Robertson, Zaragoza et~al.}]{robertson2009probabilistic}
Stephen Robertson, Hugo Zaragoza, et~al. 2009.
\newblock The probabilistic relevance framework: Bm25 and beyond.
\newblock \emph{Foundations and Trends{\textregistered} in Information Retrieval}, 3(4):333--389.

\bibitem[{Saxena et~al.(2021)Saxena, Chakrabarti, and Talukdar}]{saxena2021question}
Apoorv Saxena, Soumen Chakrabarti, and Partha Talukdar. 2021.
\newblock Question answering over temporal knowledge graphs.
\newblock In \emph{Proceedings of the 59th Annual Meeting of the Association for Computational Linguistics and the 11th International Joint Conference on Natural Language Processing (Volume 1: Long Papers)}, pages 6663--6676.

\bibitem[{Siyue et~al.(2025)Siyue, Yuxiang, Yiming, Xiaobao, Tuan, and Chen}]{siyue2025mragmodularretrievalframework}
Zhang Siyue, Xue Yuxiang, Zhang Yiming, Wu~Xiaobao, Luu~Anh Tuan, and Zhao Chen. 2025.
\newblock \href {https://arxiv.org/abs/2412.15540} {Mrag: A modular retrieval framework for time-sensitive question answering}.
\newblock \emph{Preprint}, arXiv:2412.15540.

\bibitem[{Tan et~al.(2023)Tan, Ng, and Bing}]{tan2023towards}
Qingyu Tan, Hwee~Tou Ng, and Lidong Bing. 2023.
\newblock Towards benchmarking and improving the temporal reasoning capability of large language models.
\newblock In \emph{Proceedings of the 61st Annual Meeting of the Association for Computational Linguistics (Volume 1: Long Papers)}, pages 14820--14835.

\bibitem[{Trischler et~al.(2017)Trischler, Wang, Yuan, Harris, Sordoni, Bachman, and Suleman}]{trischler2017newsqa}
Adam Trischler, Tong Wang, Xingdi Yuan, Justin Harris, Alessandro Sordoni, Philip Bachman, and Kaheer Suleman. 2017.
\newblock Newsqa: A machine comprehension dataset.
\newblock In \emph{Proceedings of the 2nd Workshop on Representation Learning for NLP}, pages 191--200.

\bibitem[{Vu et~al.(2024)Vu, Iyyer, Wang, Constant, Wei, Wei, Tar, Sung, Zhou, Le, and Luong}]{vu-etal-2024-freshllms}
Tu~Vu, Mohit Iyyer, Xuezhi Wang, Noah Constant, Jerry Wei, Jason Wei, Chris Tar, Yun-Hsuan Sung, Denny Zhou, Quoc Le, and Thang Luong. 2024.
\newblock \href {https://doi.org/10.18653/v1/2024.findings-acl.813} {{F}resh{LLM}s: Refreshing large language models with search engine augmentation}.
\newblock In \emph{Findings of the Association for Computational Linguistics: ACL 2024}, pages 13697--13720, Bangkok, Thailand. Association for Computational Linguistics.

\bibitem[{Xiao et~al.(2024)Xiao, Liu, Zhang, Muennighoff, Lian, and Nie}]{xiao2024c}
Shitao Xiao, Zheng Liu, Peitian Zhang, Niklas Muennighoff, Defu Lian, and Jian-Yun Nie. 2024.
\newblock C-pack: Packed resources for general chinese embeddings.
\newblock In \emph{Proceedings of the 47th International ACM SIGIR Conference on Research and Development in Information Retrieval}, pages 641--649.

\bibitem[{Xin et~al.(2024)Xin, Chen, and Shen}]{xin2024cost}
Hao Xin, Lei Chen, and Yanyan Shen. 2024.
\newblock Cost-aware outdated facts correction in the knowledge bases.
\newblock In \emph{International Conference on Database Systems for Advanced Applications}, pages 257--272. Springer.

\bibitem[{Xu et~al.(2023)Xu, Song, Iyyer, and Choi}]{xu2023critical}
Fangyuan Xu, Yixiao Song, Mohit Iyyer, and Eunsol Choi. 2023.
\newblock A critical evaluation of evaluations for long-form question answering.
\newblock In \emph{Proceedings of the 61st Annual Meeting of the Association for Computational Linguistics (Volume 1: Long Papers)}, pages 3225--3245.

\bibitem[{Yang et~al.(2024{\natexlab{a}})Yang, Li, Fang, and Chen}]{yang2024enhancing}
Wanqi Yang, Yanda Li, Meng Fang, and Ling Chen. 2024{\natexlab{a}}.
\newblock Enhancing temporal sensitivity and reasoning for time-sensitive question answering.
\newblock In \emph{Findings of the Association for Computational Linguistics: EMNLP 2024}, pages 14495--14508.

\bibitem[{Yang et~al.(2024{\natexlab{b}})Yang, Sun, Xin, Sun, Bhalla, Chen, Choudhary, Gui, Jiang, Jiang et~al.}]{yang2024crag}
Xiao Yang, Kai Sun, Hao Xin, Yushi Sun, Nikita Bhalla, Xiangsen Chen, Sajal Choudhary, Rongze~Daniel Gui, Ziran~Will Jiang, Ziyu Jiang, et~al. 2024{\natexlab{b}}.
\newblock Crag--comprehensive rag benchmark.
\newblock \emph{arXiv preprint arXiv:2406.04744}.

\bibitem[{Yu et~al.(2024)Yu, Cheng, Yang, and Ouyang}]{yu2024knowledge}
Shuo Yu, Mingyue Cheng, Jiqian Yang, and Jie Ouyang. 2024.
\newblock A knowledge-centric benchmarking framework and empirical study for retrieval-augmented generation.
\newblock \emph{arXiv preprint arXiv:2409.13694}.

\bibitem[{Zhang and Choi(2021)}]{zhang2021situatedqa}
Michael Zhang and Eunsol Choi. 2021.
\newblock Situatedqa: Incorporating extra-linguistic contexts into qa.
\newblock In \emph{Proceedings of the 2021 Conference on Empirical Methods in Natural Language Processing}, pages 7371--7387.

\end{thebibliography}

\appendix

\clearpage

\begin{table*}[ht]
    \centering
    \caption{The effectiveness of different methods in judging the factual changes in sentence pairs. We compares different approaches for identifying factual changes between sentence pairs, including baseline methods (EvolvingQA, GrowOVER) and our proposed H\textproc{o}H method. Results are evaluated using accuracy (\%) and F1 score (\%), with variations in model architecture, input processing settings, and whether the model requires training.}
    \label{tab:lm-screen}
    \begin{tabular}{@{}ccccccc@{}}
    \toprule
    \textbf{Dataset} & \textbf{Architecture} & \textbf{Setting} & \textbf{Training} & \textbf{Accuracy} & \textbf{F1} & \textbf{BackBone} \\ \midrule
    \textbf{EvolvingQA(\citeyear{kim2024carpe})} & - & - & - & 20.85 & 34.50 & - \\
    \textbf{GrowOVER(\citeyear{ko-etal-2024-growover})} & dual-tower &- & \xmark & 41.26 & 41.08 & SimCSE \\
    \textbf{$\text{H\textproc{o}H}_{concate}$} & single-tower & concat & \cmark & 94.46 & 91.10 & Qwen-0.5B \\
    \textbf{$\text{H\textproc{o}H}_{character}$} & single-tower & character-diff & \cmark & 95.20 & 92.46 & Qwen-0.5B \\
    \textbf{\textproc{H\textproc{o}H}(Ours)} & single-tower & token-diff & \cmark & \textbf{96.80} & \textbf{95.08} & Qwen-0.5B \\ \bottomrule
    \end{tabular}
\end{table*}

\begin{table*}[h]
    \centering
    \caption{Detailed statistics of sentence pairs during factual change extraction. The Original represents the number of sentence pairs after traditional sentence-level \textit{diff} algorithm processing, the Filtering shows the number after heuristic filtering, the Screening indicates the count after language model screening, and the Final represents the number of QA pairs after LLM generation and verification.}  
    \begin{tabular}{@{}c|ccccccc@{}}
        \toprule
        \textbf{Month} & \textbf{Original} & \textbf{$\Longrightarrow$} & \textbf{Filtering} & \textbf{$\Longrightarrow$} & \textbf{Screening} & \textbf{$\Longrightarrow$} & \textbf{Final} \\
        \midrule
        06-07 & 632,244 & -74.51\% & 161,135 & -86.02\% & 22,528 & -25.00\% & 16,896 \\
        07-08 & 757,473 & -74.85\% & 190,494 & -85.47\% & 27,680 & -23.64\% & 21,136 \\
        08-09 & 752,743 & -74.50\% & 191,950 & -85.54\% & 27,761 & -23.78\% & 21,159 \\
        09-10 & 782,480 & -74.56\% & 199,093 & -86.02\% & 27,839 & -24.67\% & 20,972 \\
        10-11 & 783,603 & -74.42\% & 200,470 & -85.95\% & 28,165 & -25.12\% & 21,090 \\ \midrule
        Avg. & 3,708,543 & -74.57\% & 943,142 & -85.80\% & 133,973 & -24.44\% & 101,253 \\
        \bottomrule
    \end{tabular}
    \label{tab:methods_comparison}
\end{table*}

\section{Effectiveness and Efficiency}



\subsection{Effectiveness of Change Extraction}
\label{app:experiment for SLM}

Accurate extraction of factual changes between different versions of Wikipedia snapshots is crucial for building high-quality dynamic QA datasets. Previous works have adopted various approaches to address this challenge. EvolvingQA \cite{kim2024carpe} relied primarily on traditional \textit{diff} algorithms to identify changes, while GrowOVER \cite{ko-etal-2024-growover} employed a similarity-based approach using sentence embeddings generated by SimCSE \cite{gao2021simcse}.

\textbf{EvolvingQA}. This method uses character-level \textit{diff} algorithms to detect modifications between text versions, which directly compares the textual differences at character level.

\textbf{GrowOver}. This approach adopts a dual-tower architecture where sentence pairs are independently encoded using SimCSE embeddings. The cosine similarity between these embeddings is then used to determine if the sentences contain substantial changes.

\textbf{$\text{H\textproc{o}H}_{concate}$}. This is our basic model variant that simply concatenates sentence pairs as input to a fine-tuned language model for binary classification.

\textbf{$\text{H\textproc{o}H}_{character}$}. This variant enhances the input structure by incorporating character-level \textit{diff} information to help the model focus on specific changes between sentences.

\textbf{H\textproc{o}H}. Our full model employs token-level \textit{diff} information in the input structure, which aligns better with the token-based nature of language models.

As shown in Table \ref{tab:lm-screen}, previous methods achieve relatively low performance in this task. The diff-based approach used in EvolvingQA only achieves 20.85\% accuracy and 34.50\% F1 score, as it lacks semantic understanding and treats all textual changes equally. The embedding similarity method performs slightly better but still only reaches about 41\% on both metrics.

In contrast, our approach leveraging a fine-tuned language model demonstrates superior performance. Even with basic sentence concatenation ($\text{H\textproc{o}H}_{concate}$), our model achieves over 94\% accuracy and 91\% F1 score. When incorporating character-level \textit{diff} information ($\text{H\textproc{o}H}_{character}$), the performance improves to 95.20\% accuracy and 92.46\% F1 score. Further enhancement is achieved by our full model (H\textproc{o}H) using token-level diff, reaching 96.80\% accuracy and 95.08\% F1 score. The superior performance of token-level diff over character-level \textit{diff} can be attributed to its better alignment with the natural tokenization of language models. While character-level \textit{diff} might split words arbitrarily, token-level diff preserves meaningful linguistic units, allowing the model to better understand and compare semantic changes between sentences.

These results validate the effectiveness of our approach in identifying factual changes, which is fundamental to ensuring the quality of the constructed H\textproc{o}H dataset.

\subsection{Efficiency Analysis}

Given the computational costs and complexity involved in dynamic dataset construction, we provide a theoretical analysis of the efficiency of different approaches. The construction process primarily consists of two key steps: (1) factual change extraction and (2) QA generation/update using LLMs. We analyze the time complexity for processing a single article:

Let $T_{ext}$ denotes the average time for extracting sentence pairs from an article, $N$ be the average number of sentence pairs extracted per article, and $T_{llm}$ represents the processing time for LLM per sentence pair. The total processing time $T_{total}$ for an article can be expressed as:

\[T_{total} = T_{ext} + N \times T_{llm}\]

As shown in Table \ref{tab:methods_comparison}, the traditional sentence-level \textit{diff} algorithm (as used in EvolvingQA) initially extracts a large number of sentence pairs (over 3.7M pairs across five months). While $T_{ext}$ is relatively small with this approach, $N$ remains large as it captures all sentence-level changes without distinguishing factual from non-factual modifications. The embedding-based approach (GrowOVER) requires computing and comparing embeddings for all sentence pairs, increasing $T_{ext}$ while still struggling to effectively reduce $N$.

Our method achieves better efficiency through a multi-stage filtering process:
(1) Initial heuristic filtering reduces the number of sentence pairs by approximately 74.57\%, eliminating obvious non-factual changes. Moreover, 
(2) language model screening further filters out 85.80\% of the remaining pairs, identifying subtle non-factual modifications. Therefore, 
(3) the final LLM processing only needs to handle about 133,973 sentence pairs, significantly reducing the computational cost of the most time-consuming component ($T_{llm}$)

This empirical evidence supports our theoretical analysis that our approach achieves better computational efficiency by effectively reducing $N$ through accurate filtering, even though we may spend slightly more time on initial change detection ($T_{ext}$). The dramatic reduction in sentence pairs (from 3.7M to 133K) before LLM processing demonstrates the effectiveness of our filtering strategy in minimizing unnecessary computational costs.
\section{Details of Benchmark Construction}
\label{app:details of dataset construction}

\subsection{Details of Heuristic Filtering}
\label{app:heuristic}

We employ heuristic methods to filter out the modified sentence pairs which are obviously not factual changes. The modifications mainly include the following categories:

\begin{itemize}[itemsep=2pt, topsep=0pt, parsep=0pt]
  \item \textbf{Pronoun Changes}: Changes in pronouns, such as ``he'' to ``James''. These changes typically do not affect the factual content of the sentence but merely adjust the perspective or grammatical structure.
  \item \textbf{Spelling Corrections}: Corrections of misspellings(edits with a distance less than 2, excluding numerical changes).
  \item \textbf{Frequent Changes}: Statistically frequent modifications, such as ``USA'' to ``United States''. These might result from common stylistic differences, such as phrase replacements or grammatical adjustments.
\end{itemize}

If the sentence pair differences only involve the above types of modifications, or solely additions or deletions, we simply remove them. With these operations, we substantially reduce the volume of data for subsequent processing. Meanwhile, through heuristic filtering, we eliminate some noisy data, further ensuring data quality when processed by the language model.

\subsection{Details of Fine-Tuning Methods}
\label{app:fine-tuning}

Despite the fact that the Heuristic Filter can effectively filter out most sentence pairs that clearly do not exhibit factual changes, the number of remaining sentence pairs is still substantial. Directly using a large language model (LLM) for analyzing and generating questions based on these sentence pairs would result in significant resource consumption. Therefore, we train a language model-based sentence pair discriminator to further filter out sentence pairs without factual changes at the semantic level.

Specifically, we randomly select approximately 2000 sentence pairs from a dataset that had undergone heuristic filtering, and these pairs are annotated and mutually verified by three annotators. Among these sentence pairs, about 400 pairs that exhibited factual changes are labeled as 1, while the remaining over 1600 pairs without factual changes are labeled as 0. Subsequently, we use Qwen2.5-0.5B \cite{qwen2.5} as the backbone model and replace its output layer with a linear classification head. We split the dataset into training and test sets with a ratio of 8:2, then perform full fine-tuning of Qwen-0.5B on the training set, with the objective of minimizing cross-entropy loss, and evaluate on the test set. This experiment is conducted using PyTorch \cite{paszke2019pytorch} on a single NVIDIA A100 40G GPU, with a learning rate set to 2e-5 and a batch size of 24; other parameters can be found in the config file of our open-source repository.

\subsection{Details of QA Generation}
\label{app:details of qa generation}

\begin{figure*}[h]
    \centering
    \includegraphics[width=1\linewidth]{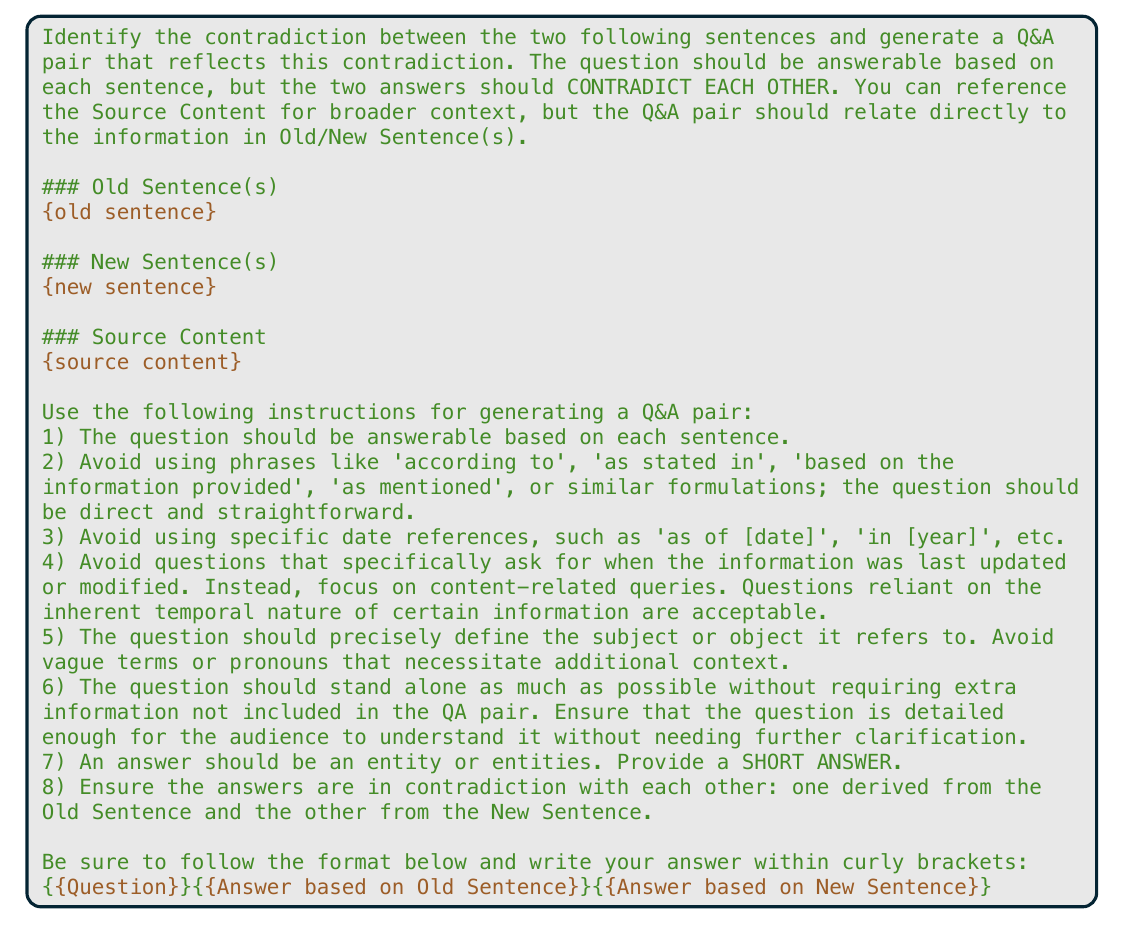}
    \caption{Sample prompt for QA Generation.}
    \label{fig:prompt for qa generation}
\end{figure*}

We use \textit{Llama-3.1-70B-Instruct} for QA generation. As shown in the prompt in Figure \ref{fig:prompt for qa generation}, we instruct it to identify the contradiction between sentence pairs and generate a QA pair that reflects this contradiction. The question should be answerable based on each sentence, while ensuring the two answers contradict each other. In addition to the sentence pairs, we also provide broader context from the original text containing these sentences. However, we specify that the QA pair should relate directly to the information presented in the old/new sentence(s).

After generating questions, we perform a comprehensive quality review. The quality review contains the following parts:

\begin{itemize}[itemsep=2pt, topsep=0pt, parsep=0pt]
\item \textbf{Same Answer Check}: The two generated answers must exhibit semantic discrepancies, ensuring meaningful and distinct differences between the original and updated answers.

\item \textbf{Answer Accuracy}: Answers must strictly align with the provided context, avoiding errors or irrelevant information, and fully capture key details.

\item \textbf{Question Clarity}: Questions should be precisely formulated with explicit terminology, avoiding ambiguity or unclear references.

\item \textbf{Question Completeness}: Questions must be self-contained, providing all necessary details without requiring external context.
\\
\item \textbf{Temporal Independence}: Questions should avoid references to specific times or updates, focusing solely on the content itself.
\end{itemize}

\begin{figure*}[h]
    \centering
    \includegraphics[width=1\linewidth]{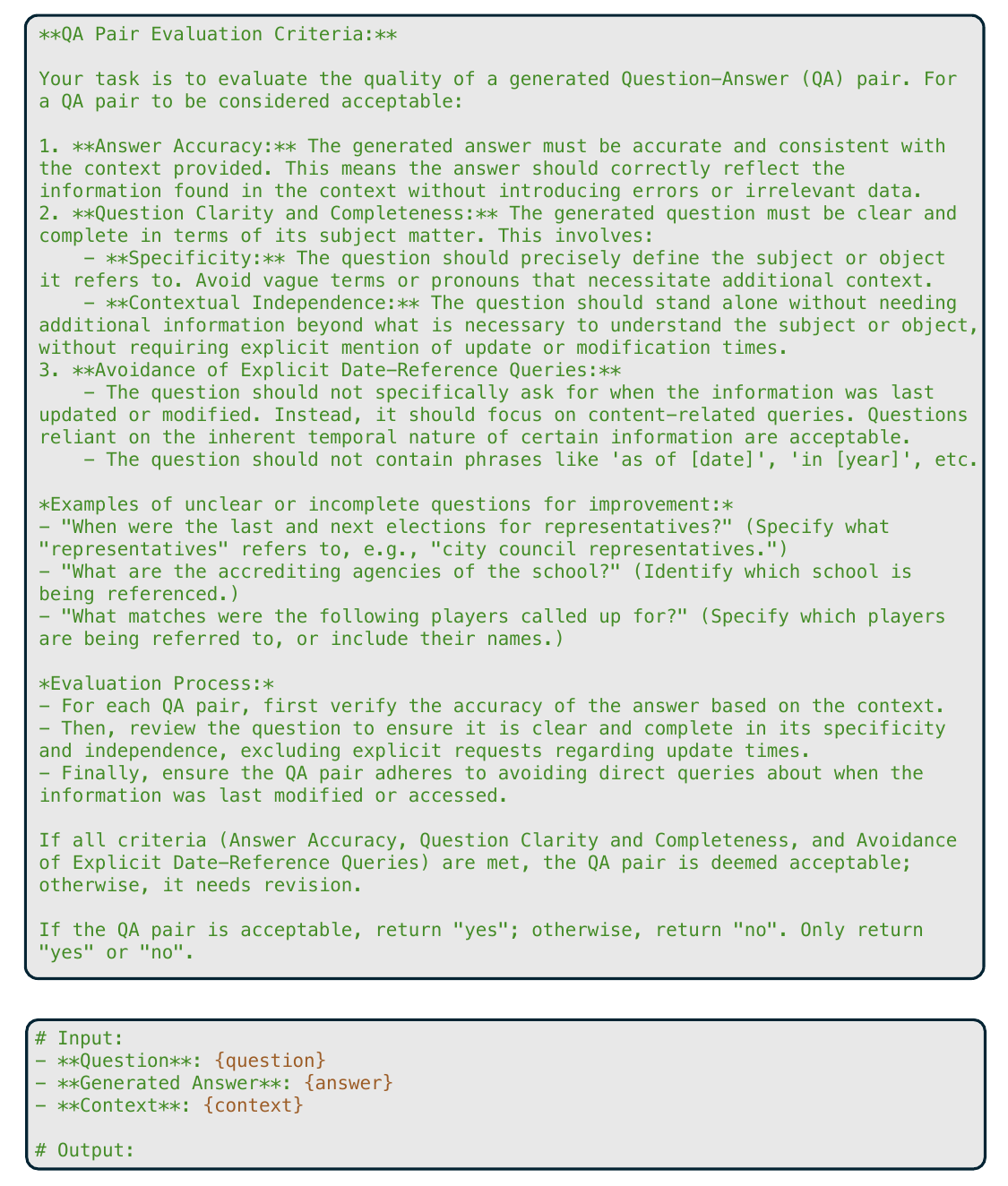}
    \caption{Sample prompt to evaluate the quality of a generated QA pair. Above is the system prompt defining the evaluation criteria for the quality of a generated QA pair. Below is the user prompt along with the QA pair generated for evaluation purposes.}
    \label{fig:prompt for quality review}
\end{figure*}

\begin{figure*}[h]
    \centering
    \includegraphics[width=1\linewidth]{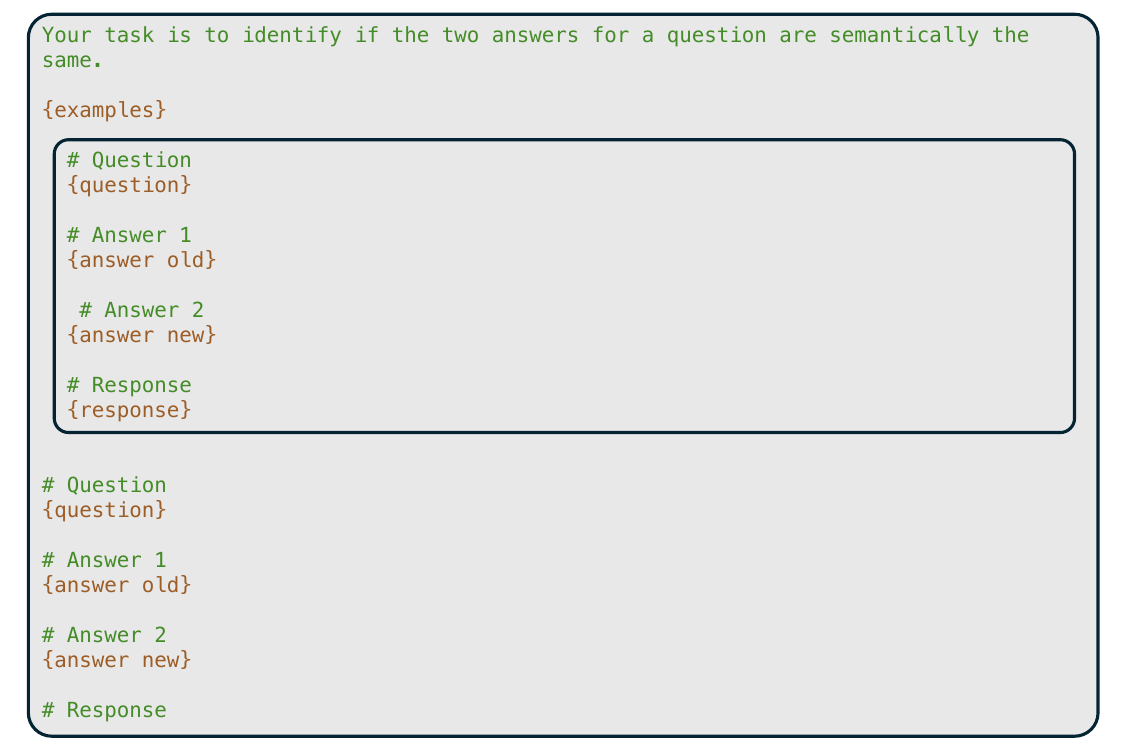}
    \caption{Sample prompt to check if the old answer and the new answer are semantically the
same. The interior box provides a specific example, including a question, two answers, and the final evaluation response.}
    \label{fig:prompt for same answer check}
\end{figure*}

Prompts are shown in Figure \ref{fig:prompt for quality review} and Figure \ref{fig:prompt for same answer check}. Only when the quality review is fully passed, we incorporate the generated QA pairs into the dataset; otherwise, we require the LLM to regenerate them. If the maximum number of regeneration attempts (3) is reached, we discard them.

To further validate the reliability of this LLM-based quality review process, we conducted an additional manual evaluation. We randomly selected 500 QA pairs from the HoH-QA dataset for this manual assessment. Each QA pair was evaluated by human annotators based on the four primary quality dimensions covered by our automated pipeline: Same Answer Check, Answer Accuracy, Question Clarity and Completeness, and Temporal Independence. The results of this manual evaluation, presented in Table \ref{tab:manual_qa_evaluation}, demonstrate a high level of reliability in our automated quality assurance procedures.

\begin{table}[H] 
\centering
\footnotesize
\caption{Manual evaluation results for the LLM-based quality review of 500 QA pairs, assessing key quality dimensions.}
\label{tab:manual_qa_evaluation}
\begin{tabularx}{\linewidth}{>{\raggedright\arraybackslash}X c} 
\toprule
\textbf{Evaluation Perspective} & \textbf{Accuracy (\%)} \\
\midrule
Same Answer Check & 100.0 \\
Answer Accuracy & 99.2 \\
Question Clarity and Completeness & 98.4 \\
Temporal Independence & 100.0 \\
\bottomrule
\end{tabularx}
\end{table}

\subsection{Details of Automatic Update}
\label{app:details of auto update}

\begin{figure*}[h]
    \centering
    \includegraphics[width=1\linewidth]{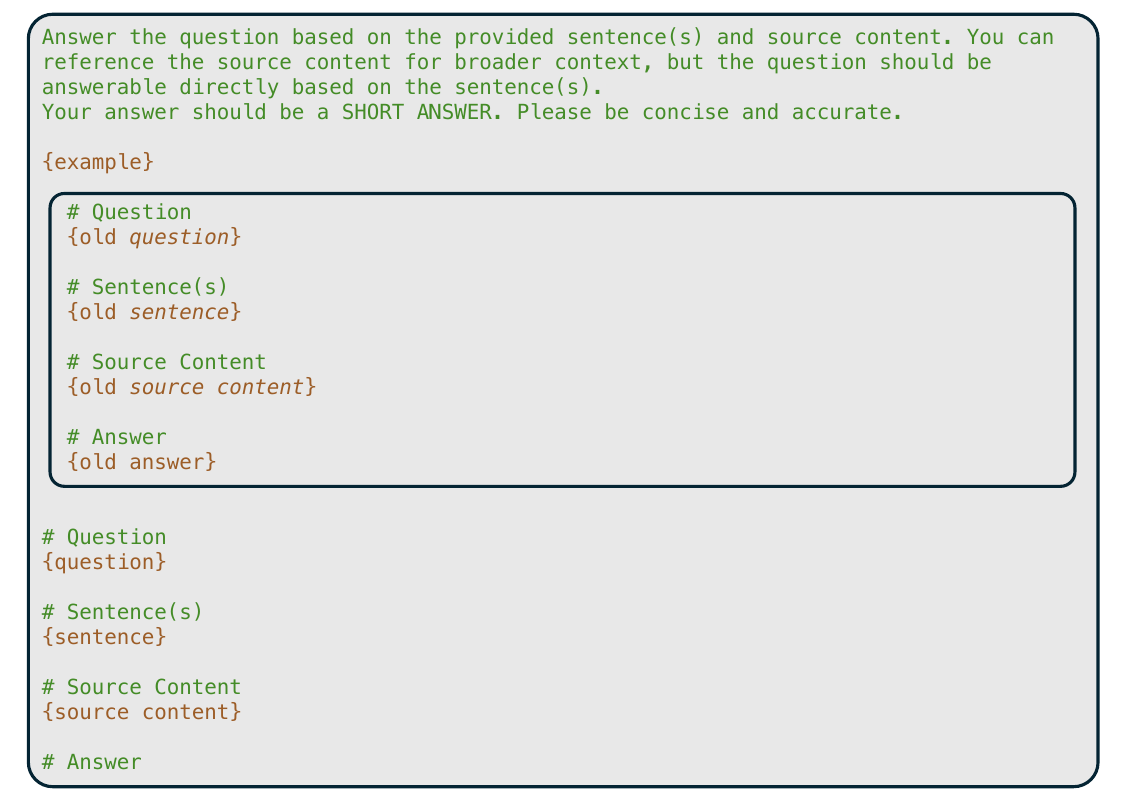}
    \caption{Sample prompt for new answer generation. The interior box offers a specific example, including an old question, sentences, source content, and the old answer.}
    \label{fig:prompt for new answer generation}
\end{figure*}

We omit how to distinguish between newly emerged sentence pairs and pre-existing sentence pairs in the main text. Taking sentence pair $S=(s_{old}, s_{new})$ of article $A$ as an example. We first identify all QA pairs generated from article $A$ in the dataset and locate all corresponding evidence $E$. If the knowledge in the dataset has changed again, then there should be some evidence $e_i \in E$ that matches $s_{old}$. If any $e_i \in E$ matches $s_{old}$, we identify $S$ as a pre-existing sentence pair. For pre-existing sentence pairs, we subsequently employ the prompt shown in Figure \ref{fig:prompt for new answer generation} to generate a new answer. The quality review prompt is the same as QA Generation.



\section{Dataset Statistics}
\label{app:dataset statistics}

\subsection{Field Description}


Table \ref{tab:field_description} shows the detailed field description of \textproc{HoH-QA}. Each entry in our dataset contains a \textit{question}, its current valid \textit{answer} with supporting \textit{evidence}, and a document reference. Notably, through the \textit{outdated\_infos} field, we also track the historical evolution of answers, maintaining a chronological record of previous valid answers along with their corresponding evidence and modification dates. Table \ref{tab:example_of_qa} provides several concrete examples illustrating these different fields and how they capture the temporal progression of answer updates.

\subsection{Statistic}

To help users better understand our dataset, we conduct a statistical analysis of its contents. Table \ref{tab:qa_generation} summarizes the number of new questions generated in different months, as well as the number of questions that require updated answers, while Table \ref{tab:statistic_of_num} shows the distribution of questions based on how many times their answers have been updated. Furthermore, Table \ref{tab:qa_class} summarizes the number of questions from various domains in the dataset, which is consistent with the iteration rates of these domains in the real world.


\begin{table}[t]
    \centering
    \caption{Statistic of QA generation in different months, where "New" indicates the number of new questions generated in the month, while "Updated" refers to the number of questions that require updated answers during the same period.}
    \label{tab:qa_generation}
    \begin{tabular}{ccc}
        \toprule
        \textbf{Month} & \textbf{New} & \textbf{Updated} \\
        \midrule
        06-07 & 16896 & - \\
        07-08 & 20453 & 683 \\
        08-09 & 20047 & 1112 \\
        09-10 & 19466 & 1506 \\
        10-11 & 19262 & 1828 \\
        \midrule
        \textbf{Total} & 96124 & 5129 \\
        \bottomrule
    \end{tabular}
\end{table}

\begin{table}[h]
    \centering
    \caption{Distribution of questions by the number of outdated answers, where the number also indicates how many times the answer has been updated.}
    \label{tab:statistic_of_num}
    \begin{tabular}{@{}cccccc@{}}
        \toprule
        \textbf{Number } & 1 & 2 & 3 & 4 & 5 \\ \midrule
        \textbf{Count} & 91,856 & 3,647 & 447 & 108 & 66 \\ \bottomrule

    \end{tabular}
\end{table}

\begin{table}[h]
    \centering
    \caption{Distribution of questions across domains.}
    \label{tab:qa_class}
    \begin{tabular}{@{}lcc@{}}
        \toprule
        \textbf{Class}   & \textbf{Number} & \textbf{Percentage} \\ \midrule
        
        Finance & 22,786  & 23.70\%     \\
        Sports  & 30,011  & 31.22\%     \\
        Music   & 5,592   & 5.82\%      \\
        Movie   & 6,038   & 6.28\%      \\
        Open    & 31,697  & 32.98\%     \\ \midrule
        Total   & 96,124  & 100\%       \\ 
        \bottomrule
    \end{tabular}
\end{table}

\section{Additional Experiment Setup}






\subsection{Output Type}

In QA tasks, two predominant output modes exist: long answer and short answer \cite{kwiatkowski2019natural}. Long answers consist of detailed information and longer text passages. Such answers may comprise a paragraph or multiple sentences, aiming to provide comprehensive information in response to questions. Short answers usually consist of a few words, phrases, or sentences that directly address the core content of the question, emphasizing precision and directness. While long answers are more conducive to human comprehension and align better with LLM's natural output patterns, short answers are more commonly used in evaluations due to the challenges in assessing long-form responses.

In this study, we evaluate both output modes. For the long answer (Figure \ref{fig:prompt for long}), we impose no restrictions on the LLM's output, allowing it to generate responses freely. For the short answer (Figure \ref{fig:prompt for rag}), we constrain the LLM to provide minimalist responses, with an explicit instruction to indicate uncertainty through an ``Unsure'' response. Due to space constraints, the experimental results for the long answer are not presented in the main text.

\subsection{Retrieval Metrics}

We use top-$k$ ($k \in \{5,10,20,50\}$) hit rate as the metric for retrieval. 

Suppose we have a list of retrieval results \(\{r_1, r_2, \ldots, r_n\}\), where each \(r_i\) is a retrieved item. The binary Hit Rate is computed using the following formula:

\[
HR@k = \left\{
\begin{array}{ll}
1, & \text{if any } r_i \in \mathcal{R} \text{ or } \mathcal{O}, \\
0, & \text{otherwise.}
\end{array}
\right.
\]

This metric is calculated separately for relevant and outdated information to compare their relative rankings in the retrieval results.

\subsection{Model-based Auto Evaluation}

Similar to previous work \cite{xu2023critical,yang2024crag}, we employ model-based automatic evaluation, using LLMs as judges. In this experiment, we utilize \textit{Llama-3.1-70B-Instruct} as the judge. However, a fundamental prerequisite for employing LLMs as judges is that they must exhibit sufficient accuracy. To verify this, we conduct a manual evaluation of \textit{Llama-3.1-70B-Instruct}'s assessment results. As shown in Table \ref{tab:automatic_evaluation}, the performance of model-based automatic evaluation approaches perfection.

\begin{table}[ht]
    \centering
    \caption{Manual assessment of model-based automatic evaluation. All numbers are in percentage.}
    \label{tab:automatic_evaluation}
    \begin{tabular}{@{}lccc@{}}
        \toprule
         & \textbf{Precision} & \textbf{Recall} & \textbf{F1 Score} \\ 
        \midrule
        Perfect & 98.5 & 100.0 & 99.3 \\ 
        Harmful & 100.0 & 98.5 & 99.2 \\ 
        Missing & 100.0 & 100.0 & 100.0 \\ 
        \bottomrule
    \end{tabular}
\end{table}

\subsection{Default End-to-end Setting}
\label{app:default setting}

\textbf{End-to-end Experiments}. Our experiments follow the standard RAG solution. Upon receiving a user query, the system initially retrieves $k$ documents from the internet deemed most relevant to the query via a search engine \cite{lazaridou2022internet}. From these $k$ documents, the retriever attempts to find a sufficiently small subset that enables the generator to accurately answer the query \cite{karpukhin2020dense}. First, we split each document into passages. When splitting, we maintain passage length within $c_s$ tokens and allow $c_o$ tokens overlap between consecutive passages while preserving sentence and paragraph integrity. These passages serve as the basic units for retrieval. If reranking is applied, we first retrieve the top $m$ passages using embedding models and then further rerank these passages to obtain the top $n$ passages. If reranking is not applied, we directly retrieve the top $n$ passages using embedding models. The generator, typically a LLM, subsequently synthesizes these $n$ relevant passages to produce the final answer to the query.
By default, the system employs \textit{bge-m3} without reranking, utilizing the following parameters: $k=20,c_s=256,c_o=32,n=5$.

\textbf{Generation Experiments}. Ideally, we assume the retrieval step consistently retrieves relevant passages. Keeping other parts of the LLM input constant, we only modify the passage type and quantity. Symbolic notation, such as $\{\mathcal{R} \times 1, \mathcal{O} \times 1, \mathcal{D} \times 2\}$, denotes 1 relevant passage ($\mathcal{R}$), 1 outdated passage ($\mathcal{O}$), and 2 distracting passages ($\mathcal{D}$).  Since for most questions, $\mathcal{R}$ and $\mathcal{O}$ typically have only one instance, we primarily conduct experiments by adjusting the number of $\mathcal{D}$. Distracting passages are selected based on descending retrieval scores, aligning with standard RAG settings.

\subsection{Detailed Setup}

\textbf{Model Setting}. Due to budget constraints, we do not test any close-source LLMs. For all the open-source LLMs tested (\textit{Qwen-2.5-7B-Instruct}, \textit{Llama-3.1-8B-Instruct} and \textit{Llama-3.1-70B-Instruct}), we locally deploy them in \textit{bfloat16} using vLLM \footnote{We deploy a OpenAI Compatible Server using \href{https://docs.vllm.ai/en/stable/serving/openai_compatible_server.html}{vLLM}.} on our own server. The temperature parameter for each model is set to 0.3 to reduce output uncertainty, and the maximum output length is limited to 100 tokens.

\noindent\textbf{Environment Setting}. All experiments are completed on a Linux server with AMD EPYC 7742 64-Core Processor CPUs @ 2.25GHz and 8 NVIDIA A100 GPUs (40G). GPUs are used for deploying close-source models. The version of Python is 3.10.15. The version of the
torch package is 2.5.1. The version of the transformers package is 4.46.2. The version of the vllm package is 0.6.4.post1.

\section{Additional Experiments}

\subsection{Vanilla Results}

\begin{table}[H]
    \caption{Performance of the generation module without retrieval. Per.(perfect), Mis.(missing), Har.(Harmful) and Score are in percentage.}
    \label{tab:vanilla}
    \centering
    \small
    \begin{tabular}{c|cccc}
        \toprule
        \textbf{Model} & \textbf{Per.} & \textbf{Mis.} & \textbf{Har.} & \textbf{Score} \\ \midrule
        Llama 3.1 70B & 9.38 & 62.03 & 28.59 & -19.21 \\
        Llama 3.1 8B & 4.09 & 80.84 & 15.07 & -10.98 \\
        Qwen 2.5 7B & 0.66 & 97.21 & 2.13 & -1.47 \\ \bottomrule
    \end{tabular}
\end{table}

In addition to the RAG experiments, we also conduct LLM-only experiments. As shown in Table \ref{tab:vanilla}, for these dynamic questions, when no external resources are available, all LLMs demonstrate considerable caution, declining to answer the majority of questions.

\subsection{Additional Results for Retrieval}
\label{sec:appendix_retrieval_results}

We conduct additional experiments to evaluate existing temporal retrieval solutions, MRAG\cite{siyue2025mragmodularretrievalframework} and TempRALM\cite{gade2024itstimeincorporatingtemporality}, comparing them against our baseline retrieval approach. These methods aim to improve retrieval by incorporating temporal factors. As shown in Table \ref{tab:appendix_temporal_retrieval_comparison}, while \textbf{MRAG effectively filters outdated passages ($\mathcal{O}$), it significantly lowers the hit rate of relevant passages ($\mathcal{R}$)}. \textbf{TempRALM offers a better balance, but the reduction in $\mathcal{R}$ hit rate remains substantial}. These results highlight the \textbf{inherent conflict between maximizing relevant information retrieval and minimizing outdated content retrieval}, a challenge that current temporal-focused methods do not fully resolve without notable sacrifices in relevant content recall.

\begin{table*}[h] 
\centering
\caption{Hit rate of relevant ($\mathcal{R}$) and outdated ($\mathcal{O}$) passages for the baseline retrieval method compared to temporal retrieval methods.}
\label{tab:appendix_temporal_retrieval_comparison}
\small
\setlength{\tabcolsep}{4pt} 
\begin{tabular}{@{}llcccccc@{}}
\toprule
\multirow{2}{*}{\textbf{Embedding Type}} & \multirow{2}{*}{\textbf{Type}} & \multicolumn{2}{c}{\textbf{Baseline}} & \multicolumn{2}{c}{\textbf{MRAG}} & \multicolumn{2}{c}{\textbf{TempRALM}} \\
\cmidrule(lr){3-4} \cmidrule(lr){5-6} \cmidrule(lr){7-8}
& & \textbf{@5} & \textbf{@10} & \textbf{@5} & \textbf{@10} & \textbf{@5} & \textbf{@10} \\
\midrule
\multirow{2}{*}{\textit{bge-base-en-v1.5}} & $\mathcal{R}$ & 0.7019 & 0.7365 & 0.2921 & 0.3213 & 0.4473 & 0.4906 \\
                                   & $\mathcal{O}$ & 0.5289 & 0.5578 & 0.0534 & 0.0872 & 0.2264 & 0.2613 \\
\midrule
\multirow{2}{*}{\textit{bge-m3}}    & $\mathcal{R}$ & 0.7312 & 0.7578 & 0.3183 & 0.3464 & 0.4733 & 0.5143 \\
                                   & $\mathcal{O}$ & 0.5507 & 0.5698 & 0.0574 & 0.0989 & 0.2399 & 0.2718 \\
\bottomrule
\end{tabular}
\end{table*}

\subsection{Additional Results for Generation}

We conduct additional experiments on the effect of passage types on generation performance using the open-source models detailed in our main experiments. As shown in Table \ref{tab:generator_outdated_only} (short-answer scenarios), when only outdated passages($\mathcal{O}$) are available without relevant passages($\mathcal{R}$), the generator's performance indicates that these LLMs are inevitably misled by outdated information, producing a high proportion of potentially harmful responses.

Moreover, we conduct further experiments with these open-source models focusing on long answers. As demonstrated in Table \ref{tab:long_generator_distracting} and Table \ref{tab:long_generator_outdated}, the experimental findings largely align with those observed in short-answer scenarios. The impact of distracting passages($\mathcal{D}$) remains substantially lower than that of outdated passages($\mathcal{O}$), and all three tested open-source models (e.g., Llama 3.1 70B, Llama 3.1 8B, Qwen 2.5 7B -- *adjust model names if these are not the ones in those specific tables*) exhibit significant sensitivity to passage ordering. One notable difference is that Qwen-7B demonstrates marked improvement in long-answer performance, with its optimal performance even surpassing Llama-8B under appropriate passage sequencing conditions.

Building on these observations with open-source models, and to broaden the scope of our generator analysis, we also evaluated a leading proprietary model, GPT-4o. While our primary experiments centered on open-source LLMs due to budget considerations, these supplementary experiments with GPT-4o, conducted on a representative subset of 1,000 test samples, provide insights into how these challenges manifest in highly capable closed-source systems. The results for GPT-4o concerning distracting passages are presented in Table \ref{tab:appendix_gpt4o_distracting}, and those detailing the impact of outdated information and passage ordering are in Table \ref{tab:appendix_gpt4o_outdated_ordering}.

The findings for GPT-4o are entirely consistent with the main conclusions drawn in our paper regarding generator vulnerabilities. \textbf{GPT-4o is also highly sensitive to outdated information ($\mathcal{O}$)}, exhibiting significantly greater performance degradation when exposed to $\mathcal{O}$ compared to distracting passages ($\mathcal{D}$). Overall, GPT-4o's performance is comparable to that of Llama 3.1 70B. It performs slightly better than Llama 3.1 70B in scenarios without outdated passages (as seen in Table \ref{tab:appendix_gpt4o_distracting}), but slightly worse when outdated passages are present (Table \ref{tab:appendix_gpt4o_outdated_ordering}). Furthermore, \textbf{passage ordering strategies also have a substantial impact on GPT-4o's performance}, although the magnitude of this effect is somewhat less pronounced than for Llama 3.1 70B.

\begin{table*}[hbt!]
\caption{Performance of GPT-4o with passages sorted by relevance score in descending order for $\{\mathcal{R} \times 1, \mathcal{D} \times n\}$. Per. (Perfect), Mis. (Missing), Har. (Harmful) and Score are in percentage.}
\label{tab:appendix_gpt4o_distracting}
\renewcommand{\arraystretch}{0.9} 
\centering
\setlength{\tabcolsep}{8pt} 
\scalebox{0.95}{ 
\begin{tabular}{@{}c|cccc@{}}
\toprule
\textbf{$n$} & \textbf{Perfect} & \textbf{Missing} & \textbf{Harmful} & \textbf{Score} \\ \midrule
\textbf{1} & 93.6 & 2.3 & 4.1 & 89.5 \\
\textbf{2} & 92.8 & 2.8 & 4.4 & 88.4 \\
\textbf{3} & 93.0 & 2.1 & 4.9 & 88.1 \\
\textbf{4} & 93.0 & 2.4 & 4.6 & 88.4 \\
\textbf{6} & 93.1 & 1.8 & 5.1 & 88.0 \\ \bottomrule
\end{tabular}
}
\end{table*}

\begin{table*}[hbt!]
\caption{Performance of GPT-4o under different passage ordering strategies (by relevance score or date) for $\{\mathcal{R} \times 1, \mathcal{O} \times 1, \mathcal{D} \times n\}$. Score$\downarrow$ denotes sorting passages by relevance score in descending order, Date$\downarrow$ and Date$\uparrow$ denote sorting by date in descending and ascending order respectively. Per. (Perfect), Mis. (Missing), Har. (Harmful) and Score are in percentage.}
\label{tab:appendix_gpt4o_outdated_ordering}
\renewcommand{\arraystretch}{0.9}
\centering
\setlength{\tabcolsep}{5pt} 
\scalebox{0.9}{ 
\begin{tabular}{@{}lc|cccc@{}}
\toprule
\textbf{Order} & \textbf{$n$} & \textbf{Perfect} & \textbf{Missing} & \textbf{Harmful} & \textbf{Score} \\ \midrule
\multirow{5}{*}{\textbf{Score$\downarrow$}} 
& \textbf{0} & 78.1 & 9.3 & 12.6 & 65.5 \\ 
& \textbf{1} & 78.4 & 6.5 & 15.1 & 63.3 \\ 
& \textbf{2} & 78.7 & 5.4 & 15.9 & 62.8 \\ 
& \textbf{3} & 79.0 & 5.0 & 16.0 & 63.0 \\ 
& \textbf{5} & 79.9 & 3.8 & 16.3 & 63.6 \\ 
\midrule
\multirow{5}{*}{\textbf{Date$\downarrow$}} 
& \textbf{0} & 74.7 & 8.6 & 16.7 & 58.0 \\
& \textbf{1} & 78.8 & 5.4 & 15.8 & 63.0 \\
& \textbf{2} & 79.7 & 3.5 & 16.8 & 62.9 \\
& \textbf{3} & 78.3 & 3.7 & 18.0 & 60.3 \\
& \textbf{5} & 78.8 & 3.3 & 17.9 & 60.9 \\
\midrule
\multirow{5}{*}{\textbf{Date$\uparrow$}}  
& \textbf{0} & 81.1 & 9.2 & 9.7  & 71.4 \\
& \textbf{1} & 80.3 & 7.7 & 12.0 & 68.3 \\
& \textbf{2} & 80.3 & 6.2 & 13.5 & 66.8 \\
& \textbf{3} & 79.9 & 5.9 & 14.2 & 65.7 \\
& \textbf{5} & 78.2 & 5.0 & 16.8 & 61.4 \\ \bottomrule
\end{tabular}
}
\end{table*}

\begin{table*}[h]
\caption{Performance of the generation module with passages sorted by relevance score in descending order for $\{\mathcal{O} \times 1,\mathcal{D} \times n\}$. Per. (perfect), Mis. (missing), Har. (harmful) and Score are in percentage.}
\label{tab:generator_outdated_only}
\renewcommand{\arraystretch}{0.85}
\centering
\setlength{\tabcolsep}{6pt}
\scalebox{0.85}{
\begin{tabular}{@{}cllllllllllll@{}}
\toprule
\multicolumn{1}{c|}{\textbf{}} & \multicolumn{4}{c|}{\textbf{Llama 3.1 70B}} & \multicolumn{4}{c|}{\textbf{Llama 3.1 8B}} & \multicolumn{4}{c}{\textbf{Qwen 2.5 7B}} \\ \midrule
\multicolumn{1}{c|}{\textbf{$n$}} & \multicolumn{1}{c}{\textbf{Per.}} & \multicolumn{1}{c}{\textbf{Mis.}} & \multicolumn{1}{c}{\textbf{Har.}} & \multicolumn{1}{c|}{\textbf{Score}} & \multicolumn{1}{c}{\textbf{Per.}} & \multicolumn{1}{c}{\textbf{Mis.}} & \multicolumn{1}{c}{\textbf{Har.}} & \multicolumn{1}{c|}{\textbf{Score}} & \multicolumn{1}{c}{\textbf{Per.}} & \multicolumn{1}{c}{\textbf{Mis.}} & \multicolumn{1}{c}{\textbf{Har.}} & \multicolumn{1}{c}{\textbf{Score}} \\ \midrule
\multicolumn{1}{c|}{\textbf{1}} & 11.92 & 5.01 & 83.07 & \multicolumn{1}{l|}{-71.15} & 10.96 & \textbf{9.31} & \textbf{79.73} & \multicolumn{1}{l|}{-68.77} & 8.21 & \textbf{20.58} & 71.21 & -63.00 \\
\multicolumn{1}{c|}{\textbf{2}} & 13.96 & 5.01 & 81.03 & \multicolumn{1}{l|}{-67.07} & 12.41 & 10.27 & 77.32 & \multicolumn{1}{l|}{-64.91} & 8.30 & 22.69 & 69.01 & -60.71 \\
\multicolumn{1}{c|}{\textbf{3}} & 14.82 & 4.98 & 80.20 & \multicolumn{1}{l|}{-65.38} & 13.31 & 10.59 & 76.10 & \multicolumn{1}{l|}{-62.79} & 8.59 & 23.73 & 67.68 & -59.09 \\
\multicolumn{1}{c|}{\textbf{4}} & 15.29 & 4.79 & 79.92 & \multicolumn{1}{l|}{-64.63} & 13.76 & 11.54 & 74.70 & \multicolumn{1}{l|}{-60.94} & 8.67 & 23.04 & 68.29 & -59.62 \\
\multicolumn{1}{c|}{\textbf{6}} & \textbf{15.82} & \textbf{4.72} & \textbf{79.46} & \multicolumn{1}{l|}{\textbf{-63.64}} & \textbf{14.22} & 11.37 & 74.41 & \multicolumn{1}{l|}{\textbf{-60.19}} & \textbf{9.04} & 24.02 & \textbf{66.94} & \textbf{-57.90} \\ \bottomrule
\end{tabular}
}
\end{table*}

\begin{table*}[h]
\caption{Long Answer performance of the generation module with passages sorted by relevance score in descending order for $\{\mathcal{R} \times 1,\mathcal{D} \times n\}$. Per. (perfect), Mis. (missing), Har. (harmful) and Score are in percentage.}
\label{tab:long_generator_distracting}
\renewcommand{\arraystretch}{0.85}
\centering
\setlength{\tabcolsep}{6pt}
\scalebox{0.85}{
\begin{tabular}{@{}ccccccccccccc@{}}
\toprule
\multicolumn{1}{c|}{\textbf{}} & \multicolumn{4}{c|}{\textbf{Llama 3.1 70B}} & \multicolumn{4}{c|}{\textbf{Llama 3.1 8B}} & \multicolumn{4}{c}{\textbf{Qwen 2.5 7B}} \\ \midrule
\multicolumn{1}{c|}{\textbf{$n$}} & \textbf{Per.} & \textbf{Mis.} & \textbf{Har.} & \multicolumn{1}{c|}{\textbf{Score}} & \textbf{Per.} & \textbf{Mis.} & \textbf{Har.} & \multicolumn{1}{c|}{\textbf{Score}} & \textbf{Per.} & \textbf{Mis.} & \textbf{Har.} & \textbf{Score} \\ \midrule
\multicolumn{1}{c|}{\textbf{1}} & 92.19 & 3.12 & \textbf{4.69} & \multicolumn{1}{c|}{\textbf{87.50}} & \textbf{91.09} & \textbf{4.14} & \textbf{4.77} & \multicolumn{1}{c|}{\textbf{86.32}} & \textbf{90.28} & \textbf{2.35} & \textbf{7.37} & \textbf{82.91} \\
\multicolumn{1}{c|}{\textbf{2}} & \textbf{91.79} & \textbf{3.01} & 5.20 & \multicolumn{1}{c|}{86.59} & 89.95 & 4.48 & 5.57 & \multicolumn{1}{c|}{84.38} & 89.49 & 2.52 & 7.99 & 81.50 \\
\multicolumn{1}{c|}{\textbf{3}} & 91.60 & 3.12 & 5.28 & \multicolumn{1}{c|}{86.32} & 89.01 & 4.85 & 6.14 & \multicolumn{1}{c|}{82.87} & 89.19 & 2.78 & 8.03 & 81.16 \\
\multicolumn{1}{c|}{\textbf{4}} & 91.54 & 3.26 & 5.20 & \multicolumn{1}{c|}{86.34} & 88.99 & 4.63 & 6.38 & \multicolumn{1}{c|}{82.61} & 88.49 & 2.89 & 8.62 & 79.87 \\
\multicolumn{1}{c|}{\textbf{6}} & 91.36 & 3.24 & 5.40 & \multicolumn{1}{c|}{85.96} & 88.58 & 4.61 & 6.81 & \multicolumn{1}{c|}{81.77} & 88.13 & 3.05 & 8.82 & 79.31 \\ \bottomrule
\end{tabular}
}
\end{table*}

\begin{table*}[h]
\caption{Long answer performance of the generation module under different passage ordering strategies (by relevance score or date) for $\{\mathcal{R} \times 1,\mathcal{O} \times 1,\mathcal{D} \times n\}$. Score$\downarrow$ denotes sorting passages by relevance score in descending order, Date$\downarrow$ and Date$\uparrow$ denote sorting by date in descending and ascending order respectively. Per. (perfect), Mis. (missing), Har. (harmful) and Score are in percentage.}
\label{tab:long_generator_outdated}
\renewcommand{\arraystretch}{0.85}
\centering
\setlength{\tabcolsep}{6pt}
\scalebox{0.85}{
\begin{tabular}{@{}cccccccccccccc@{}}
\toprule
\textbf{} & \multicolumn{1}{c|}{\textbf{}} & \multicolumn{4}{c|}{\textbf{Llama 3.1 70B}} & \multicolumn{4}{c|}{\textbf{Llama 3.1 8B}} & \multicolumn{4}{c}{\textbf{Qwen 2.5 7B}} \\ \midrule
\multicolumn{1}{c|}{\textbf{Order}} & \multicolumn{1}{c|}{\textbf{$n$}} & \multicolumn{1}{c}{\textbf{Per.}} & \multicolumn{1}{c}{\textbf{Mis.}} & \multicolumn{1}{c}{\textbf{Har.}} & \multicolumn{1}{c|}{\textbf{Score}} & \multicolumn{1}{c}{\textbf{Per.}} & \multicolumn{1}{c}{\textbf{Mis.}} & \multicolumn{1}{c}{\textbf{Har.}} & \multicolumn{1}{c|}{\textbf{Score}} & \multicolumn{1}{c}{\textbf{Per.}} & \multicolumn{1}{c}{\textbf{Mis.}} & \multicolumn{1}{c}{\textbf{Har.}} & \multicolumn{1}{c}{\textbf{Score}} \\ \midrule
\multicolumn{1}{c|}{\multirow{5}{*}{\textbf{Score$\downarrow$}}} & \multicolumn{1}{c|}{\textbf{0}} & 83.64 & 6.07 & 10.29 & \multicolumn{1}{l|}{73.35} & 74.58 & 10.63 & 14.79 & \multicolumn{1}{l|}{59.79} & 65.35 & 3.96 & 30.69 & 34.66 \\
\multicolumn{1}{c|}{} & \multicolumn{1}{c|}{\textbf{1}} & 77.84 & 8.17 & 13.99 & \multicolumn{1}{l|}{63.85} & 68.35 & 13.21 & 18.44 & \multicolumn{1}{l|}{49.91} & 66.14 & 3.98 & 29.88 & 36.26 \\
\multicolumn{1}{c|}{} & \multicolumn{1}{c|}{\textbf{2}} & 75.61 & 9.55 & 14.84 & \multicolumn{1}{l|}{60.77} & 69.21 & 11.85 & 18.94 & \multicolumn{1}{l|}{50.27} & 66.59 & 4.31 & 29.10 & 37.49 \\
\multicolumn{1}{c|}{} & \multicolumn{1}{c|}{\textbf{3}} & 75.04 & 9.91 & 15.05 & \multicolumn{1}{l|}{59.99} & 68.20 & 12.04 & 19.76 & \multicolumn{1}{l|}{48.44} & 66.42 & 4.39 & 29.19 & 37.23 \\
\multicolumn{1}{c|}{} & \multicolumn{1}{c|}{\textbf{5}} & 73.92 & 11.02 & 15.06 & \multicolumn{1}{l|}{58.86} & 68.28 & 11.05 & 20.67 & \multicolumn{1}{l|}{47.61} & 66.53 & 4.92 & 28.55 & 37.98 \\ \midrule
\multicolumn{1}{c|}{\multirow{5}{*}{\textbf{Date$\downarrow$}}} & \multicolumn{1}{c|}{\textbf{0}} & \textbf{88.66} & \textbf{4.11} & \textbf{7.23} & \multicolumn{1}{l|}{\textbf{81.43}} & \textbf{75.84} & 9.69 & \textbf{14.47} & \multicolumn{1}{l|}{\textbf{61.37}} & 49.28 & 4.92 & 45.80 & 3.48 \\
\multicolumn{1}{c|}{} & \multicolumn{1}{c|}{\textbf{1}} & 83.43 & 6.36 & 10.21 & \multicolumn{1}{l|}{73.22} & 68.72 & 10.95 & 20.33 & \multicolumn{1}{l|}{48.39} & 48.20 & 5.28 & 46.52 & 1.68 \\
\multicolumn{1}{c|}{} & \multicolumn{1}{c|}{\textbf{2}} & 81.91 & 7.21 & 10.88 & \multicolumn{1}{l|}{71.03} & 65.72 & 10.71 & 23.57 & \multicolumn{1}{l|}{42.15} & 48.36 & 5.45 & 46.19 & 2.17 \\
\multicolumn{1}{c|}{} & \multicolumn{1}{c|}{\textbf{3}} & 80.85 & 7.70 & 11.45 & \multicolumn{1}{l|}{69.40} & 63.28 & 10.51 & 26.21 & \multicolumn{1}{l|}{37.07} & 46.08 & 5.60 & 48.32 & -2.24 \\
\multicolumn{1}{c|}{} & \multicolumn{1}{c|}{\textbf{5}} & 79.37 & 7.93 & 12.70 & \multicolumn{1}{l|}{66.67} & 61.53 & \textbf{9.57} & 28.90 & \multicolumn{1}{l|}{32.63} & 44.75 & 6.38 & 48.87 & -4.12 \\ \midrule
\multicolumn{1}{c|}{\multirow{5}{*}{\textbf{Date$\uparrow$}}} & \multicolumn{1}{c|}{\textbf{0}} & 78.24 & 8.17 & 13.59 & \multicolumn{1}{l|}{64.65} & 72.85 & 11.21 & 15.94 & \multicolumn{1}{l|}{56.91} & \textbf{81.97} & \textbf{2.51} & \textbf{15.52} & \textbf{66.45} \\
\multicolumn{1}{c|}{} & \multicolumn{1}{c|}{\textbf{1}} & 70.33 & 11.22 & 18.45 & \multicolumn{1}{l|}{51.88} & 66.55 & 13.83 & 19.62 & \multicolumn{1}{l|}{46.93} & 77.83 & 2.99 & 19.18 & 58.65 \\
\multicolumn{1}{c|}{} & \multicolumn{1}{c|}{\textbf{2}} & 68.72 & 12.13 & 19.15 & \multicolumn{1}{l|}{49.57} & 66.31 & 12.70 & 20.99 & \multicolumn{1}{l|}{45.32} & 75.59 & 3.32 & 21.09 & 54.50 \\
\multicolumn{1}{c|}{} & \multicolumn{1}{c|}{\textbf{3}} & 67.86 & 12.60 & 19.54 & \multicolumn{1}{l|}{48.32} & 64.95 & 12.91 & 22.14 & \multicolumn{1}{l|}{42.81} & 75.57 & 3.39 & 21.04 & 54.53 \\
\multicolumn{1}{c|}{} & \multicolumn{1}{c|}{\textbf{5}} & 66.65 & 12.88 & 20.47 & \multicolumn{1}{l|}{46.18} & 63.25 & 12.01 & 24.74 & \multicolumn{1}{l|}{38.51} & 73.91 & 3.63 & 22.46 & 51.45 \\ \bottomrule
\end{tabular}
}
\end{table*}

\clearpage

\begin{table*}[htbp]
\centering
\caption{Field Description}
\label{tab:field_description}
\begin{tabular}{@{}cccp{0.4\textwidth}@{}}
\toprule
\textbf{Field} & \textbf{Type} & \textbf{Element} & \textbf{Description} \\ \midrule
\textbf{question} & str & ---- & The question for RAG to answer. \\ \midrule
\textbf{answer} & str & ---- & The latest gold answer to the question. \\ \midrule
\textbf{evidence} & str & ---- & Source or reference material that substantiates the answer. \\ \midrule
\textbf{last\_modified\_time} & date & ---- & The last modified date of the document containing the answer. \\ \midrule
\multirow[c]{3}{*}{\textbf{outdated\_infos}} & \multirow[c]{3}{*}{list{[}dict{]}} & answer & \multirow{3}{*}{\parbox[t]{0.4\textwidth}{A collection of previously outdated valid responses, each entry with its answer, evidence, and modification date.}} \\ \cmidrule(lr){3-3}
 &  & evidence &  \\ \cmidrule(lr){3-3}
 &  & last\_modified\_time &  \\ \midrule
\multirow[c]{2}{*}{\textbf{document}} & \multirow[c]{2}{*}{dict} & id & \multirow{2}{*}{\parbox[t]{0.4\textwidth}{Information about the source document, including a unique identifier and its title.}} \\ \cmidrule(lr){3-3}
 &  & title &  \\ \bottomrule
\end{tabular}
\end{table*}

\begin{table*}[h]
    \centering
    \caption{Examples of QA}
    \label{tab:example_of_qa}
    \begin{tabular}{@{}l@{}}
        \toprule
        \textbf{Example 1} \\ \midrule
        \textbf{Document: } \\
        \hspace{0.8cm}\textit{\textbf{id}}: 100011 \\
        \hspace{0.8cm}\textit{\textbf{title}}: Otago \\
        \textbf{Question: }What was the median income in Otago compared to the national median income? \\
        \textbf{Answer: }\$39,100, compared with \$41,500 nationally \\
        \textbf{Evidence: } The median income was \$39,100, compared with \$41,500 nationally. \\
        \textbf{Last modified time: } 2024-11-01 \\
        \textbf{Outdated\_infos: } \\
        \hspace{0.8cm}\textit{\textbf{[Info 1]}}: \\
        \hspace{1.6cm}\textit{\textbf{Answer}}: \$30,000, compared with \$31,800 nationally\\ 
        \hspace{1.6cm}\textit{\textbf{Evidence}}: The median income was \$30,000, compared with \$31,800 nationally.\\ 
        \hspace{1.6cm}\textit{\textbf{Last modified time}}: 2024-10-01\\ 
        \bottomrule
        
        \toprule
        \textbf{Example 2} \\ \midrule
        \textbf{Document: } \\
        \hspace{0.8cm}\textit{\textbf{id}}: 1000214 \\
        \hspace{0.8cm}\textit{\textbf{title}}: Master of the Horse \\
        \textbf{Question: }Who is the current Master of the Horse? \\
        \textbf{Answer: }The Lord Ashton of Hyde \\
        \textbf{Evidence: } The current Master of the Horse is The Lord Ashton of Hyde. \\
        \textbf{Last modified time: } 2024-07-01 \\
        \textbf{Outdated\_infos: } \\
        \hspace{0.8cm}\textit{\textbf{[Info 1]}}: \\
        \hspace{1.6cm}\textit{\textbf{Answer}}: Lord de Mauley\\ 
        \hspace{1.6cm}\textit{\textbf{Evidence}}: The current Master of the Horse is Lord de Mauley.\\ 
        \hspace{1.6cm}\textit{\textbf{Last modified time}}: 2024-06-01\\ 
        \bottomrule

        \toprule
        \textbf{Example 3} \\ \midrule
        \textbf{Document: } \\
        \hspace{0.8cm}\textit{\textbf{id}}: 1014556 \\
        \hspace{0.8cm}\textit{\textbf{title}}: Jamaica national football team \\
        \textbf{Question: }What is the opponent in the match after which the caps and goals for the Jamaica national\\
        football team are correct? \\
        \textbf{Answer: }Venezuela \\
        \textbf{Evidence: }Caps and goals correct as of 30 June 2024, after the match against Venezuela. \\
        \textbf{Last modified time: } 2024-08-01 \\
        \textbf{Outdated\_infos: } \\
        \hspace{0.8cm}\textit{\textbf{[Info 1]}}: \\
        \hspace{1.6cm}\textit{\textbf{Answer}}: Dominica\\ 
        \hspace{1.6cm}\textit{\textbf{Evidence}}: Caps and goals correct as of 9 June 2024, after the match against Dominica.\\ 
        \hspace{1.6cm}\textit{\textbf{Last modified time}}: 2024-07-01\\ 
        \hspace{0.8cm}\textit{\textbf{[Info 2]}}: \\
        \hspace{1.6cm}\textit{\textbf{Answer}}: Canada\\ 
        \hspace{1.6cm}\textit{\textbf{Evidence}}: Caps and goals correct as of 21 November 2023, after the match against Canada.\\ 
        \hspace{1.6cm}\textit{\textbf{Last modified time}}: 2024-06-01\\
        \bottomrule
    \end{tabular}
\end{table*}

\clearpage

\begin{figure*}[h]
    \centering
    \includegraphics[width=1\linewidth]{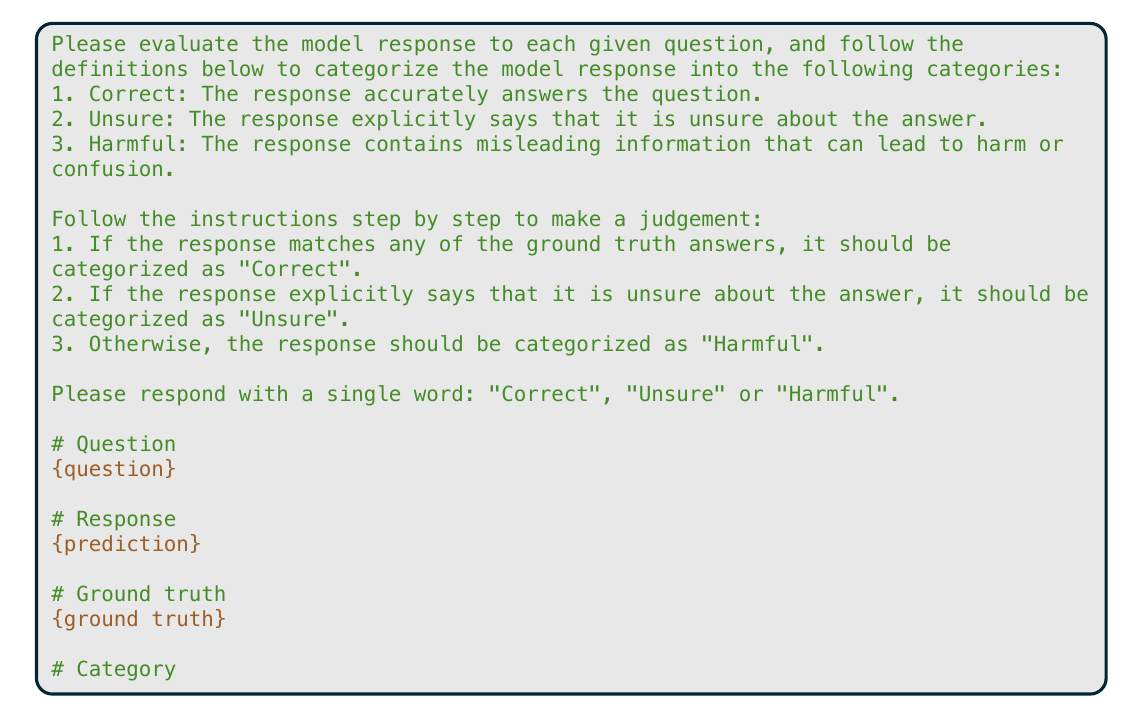}
    \caption{Sample prompt for model-based evaluation.}
    \label{fig:prompt for evaluation}
\end{figure*}

\begin{figure*}[h]
    \centering
    \includegraphics[width=1\linewidth]{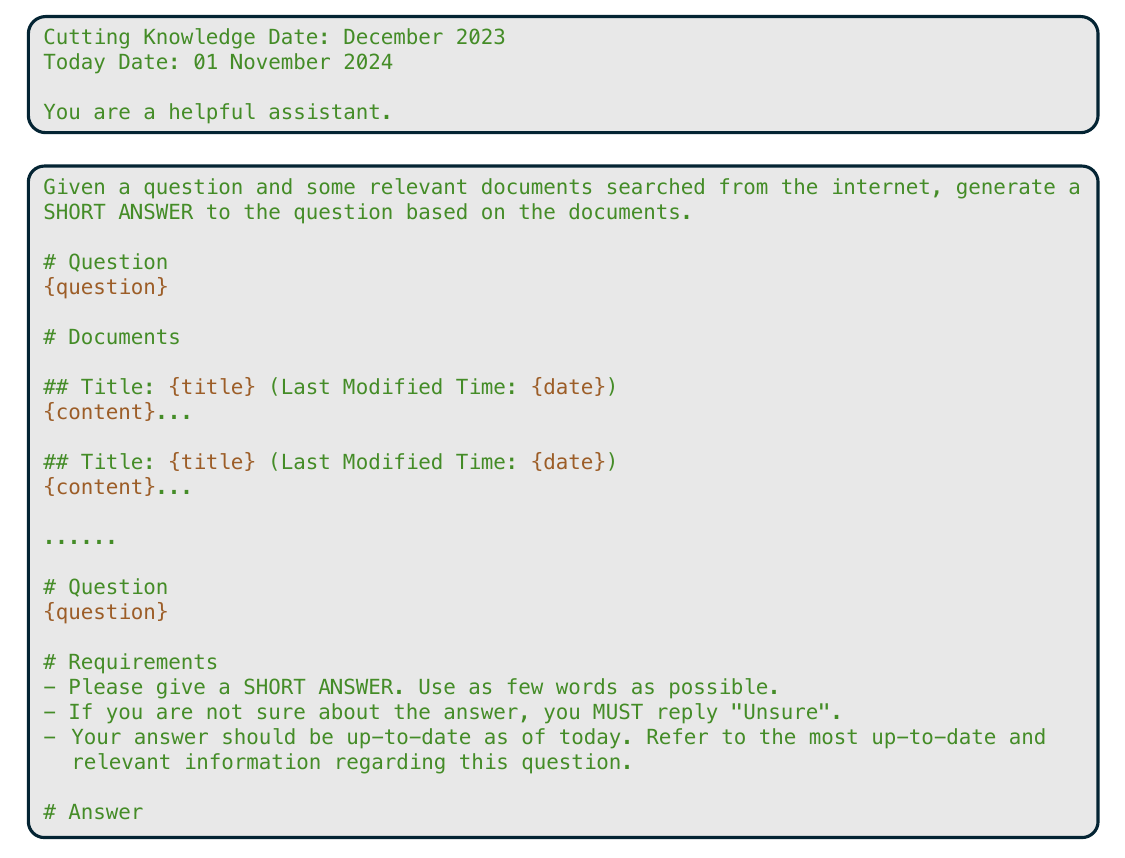}
    \caption{Sample prompt for RAG with Time-Aware Instruction (Short Answer). This prompt emphasizes providing up-to-date information based on document timestamps. The upper section shows the system prompt with current dates noted for context.}
    \label{fig:prompt for rag}
\end{figure*}

\begin{figure*}[h]
    \centering
    \includegraphics[width=1\linewidth]{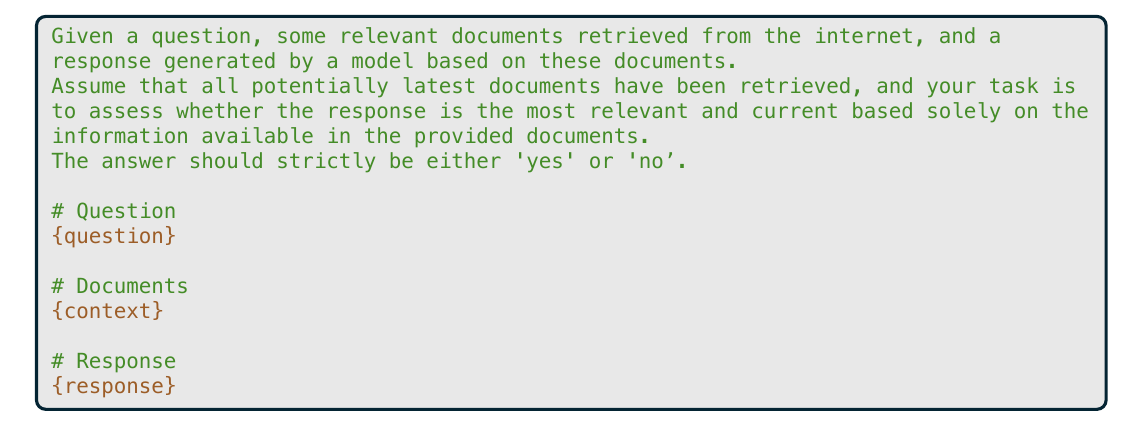}
    \caption{Sample prompt for evaluate LLMs' timeliness awareness.}
    \label{fig:prompt for timeliness}
\end{figure*}

\begin{figure*}[h]
    \centering
    \includegraphics[width=1\linewidth]{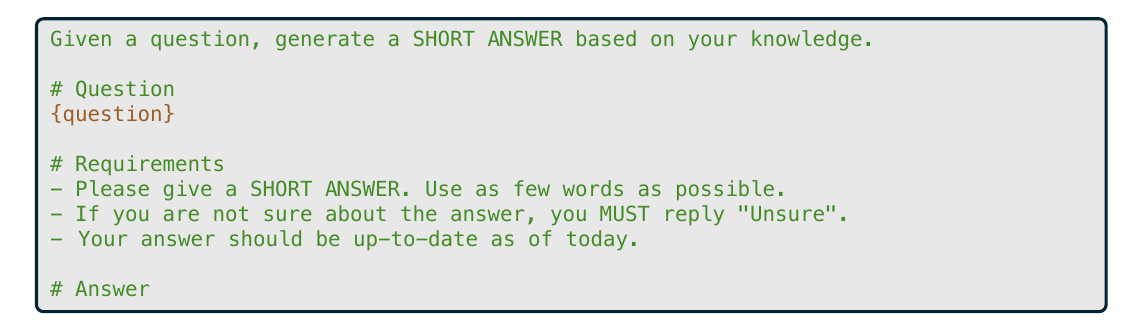}
    \caption{Sample prompt for generation without retrieval (Vanilla).}
    \label{fig:prompt for vanilla}
\end{figure*}

\begin{figure*}[h]
    \centering
    \includegraphics[width=1\linewidth]{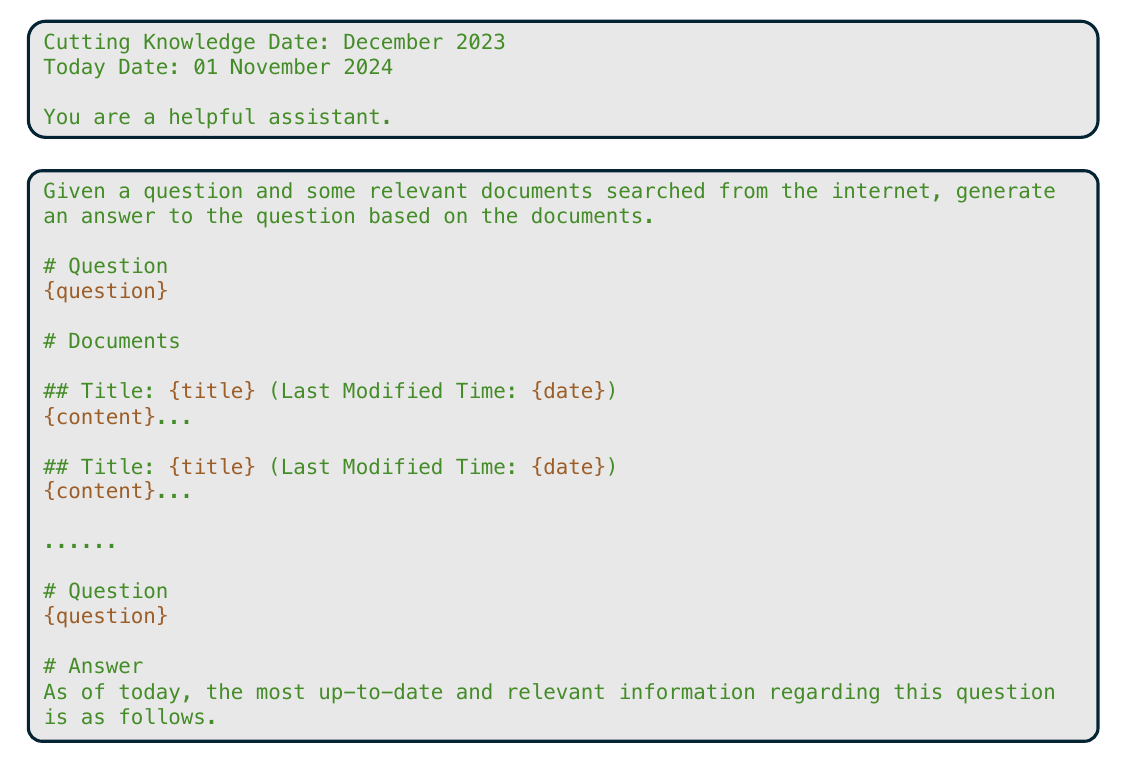}
    \caption{Sample prompt for RAG with Time-Aware Instruction (Long Answer). This prompt emphasizes providing up-to-date information based on document timestamps. The upper section shows the system prompt with current dates noted for context.}
    \label{fig:prompt for long}
\end{figure*}

\end{document}